\documentclass[12pt]{article}

\usepackage[margin=1in]{geometry}
\usepackage[utf8]{inputenc}
\usepackage{amsmath, mathtools, amsfonts, amssymb, amsthm, bm, float, subcaption}
\usepackage[shortlabels]{enumitem}
\usepackage[linesnumbered, ruled]{algorithm2e}
\usepackage{color}
\usepackage[title]{appendix}
\usepackage{tikz,wrapfig}
\usetikzlibrary{positioning}
\usepackage{hyperref}
\usepackage{url}
\usepackage{multicol}
\usepackage{setspace}
\usepackage{verbatim}
\usepackage{booktabs}
\usepackage{multirow}

\usepackage[normalem]{ulem}   % for \sout
\usepackage{xcolor}           % for \textcolor

\usepackage{tabularx}
\usepackage{makecell}

\newtheorem{theorem}{Theorem}
\newtheorem{proposition}{Proposition}
\newtheorem{lemma}{Lemma}
\newtheorem{definition}{Definition}
\newtheorem{assumption}{Assumption}

\newcommand{\cF}{\mathcal{F}}
\newcommand{\cM}{\mathcal{M}}
\newcommand{\cE}{\mathcal{E}}
\newcommand{\cD}{\mathcal{D}}
\newcommand{\cP}{\mathcal{P}}
\newcommand{\cB}{\mathcal{B}}
\newcommand{\cA}{\mathcal{A}}
\newcommand{\cC}{\mathcal{C}}
\newcommand{\frakF}{\mathfrak{F}_\tau}
\newcommand{\Lip}{{\rm Lip}}
\newcommand{\RR}{\mathbb{R}}
\newcommand{\bz}{\boldsymbol{z}}
\newcommand{\bc}{\boldsymbol{c}}
\newcommand{\bu}{\boldsymbol{u}}
\newcommand{\bx}{\boldsymbol{x}}
\newcommand{\bb}{\boldsymbol{b}}
\newcommand{\bbf}{\boldsymbol{f}}

\newcommand{\commentout}[1]{}
\newcommand{\bzeta}{\boldsymbol{\zeta}}

\usepackage[authoryear,longnamesfirst]{natbib}

\begin{document}

% Manual arXiv-style title block
\begin{center}

{\Large\bfseries
Latent Evolution Operator (LEO):\\
State-to-State Forecasting of Parametric PDEs
}

\vspace{1.0em}

Khalid Rafiq$^{1}$ \quad
Wenjing Liao$^{2}$ \quad
Aditya G. Nair$^{1,\ast}$

\vspace{0.7em}

{\small
$^{1}$Department of Mechanical Engineering, University of Nevada, Reno,\\
1664 N. Virginia St., Reno, NV 89557, USA
}

\vspace{0.25em}

{\small
$^{2}$School of Mathematics, Georgia Institute of Technology, Atlanta, GA, USA
}

\vspace{0.5em}

{\small
\texttt{krafiq@unr.edu}, \texttt{wliao60@gatech.edu}, \texttt{adityan@unr.edu}
}

\vspace{0.25em}

{\small
$^\ast$Corresponding author
}

\end{center}

\vspace{1.5em}

% Here goes the abstract
\begin{abstract}
Solving time-dependent parametric partial differential equations (PDEs) with high-fidelity solvers is expensive in multi-query settings. This motivates reduced-order models in latent space. Existing methods are typically either (i) recurrent latent dynamics models, which require step-by-step integration and accumulate error over long horizons, or (ii) neural operators, which enable direct prediction but are usually trained from initial conditions and do not explicitly enforce generalization from intermediate states.

We introduce the Latent Evolution Operator (LEO), a parameterized map that takes a state at an arbitrary time, together with physical parameters and a forecast horizon, and predicts the future state in a single latent-space evaluation. Under standard manifold and regularity assumptions, finite-time evolution operators admit an encoder--propagator--decoder decomposition. We approximate this structure using neural networks trained on state pairs sampled across anchor times and horizons.

We evaluate the method on the 1D Burgers' equation, a 2D advection--diffusion system, and the shallow-water equations. The model produces accurate predictions across parameters and time horizons, achieves significant speedups compared to Autoencoder--LSTM baselines, and maintains accuracy when initialized from intermediate states. These results indicate that learning finite-time evolution operators in latent space provides an effective alternative to recurrent latent dynamics models and operator learning methods trained only from initial conditions.
\end{abstract}


%\nocite{*}

% Main text
\section{Introduction}
The computational solution of time-dependent parametric partial differential equations (PDEs)
is a cornerstone of scientific simulation in fluid mechanics, materials science, climate modeling,
and biological systems.
For systems governed by
\begin{equation}
\frac{\partial \boldsymbol{u}}{\partial t}
  = \boldsymbol{F}\bigl(\boldsymbol{x}, \boldsymbol{u},
           \nabla_{\boldsymbol{x}} \boldsymbol{u},
           \nabla^2_{\boldsymbol{x}} \boldsymbol{u}, \ldots;\,\boldsymbol{\zeta}\bigr),
\label{eq:parametric_pde}
\end{equation}
where $\boldsymbol{u}(\boldsymbol{x},t)$ denotes the state field and $\boldsymbol{\zeta}$
collects physical, geometric, or control parameters.
High-fidelity numerical solvers quickly become prohibitively expensive in multi-query
settings such as design optimization, uncertainty quantification, inverse problems,
and real-time decision-making.
Even with modern high-performance computing, repeated forward simulations across
time horizons and parameter variations often dominate computational cost.

Our objective is to approximate the finite-time solution operator of parametric PDEs directly, mapping a system state at time $t$ to a future state at time $t+\tau$ without relying on temporal discretization or recursive rollout.
In this work, we propose to learn this operator in a low-dimensional latent space.

\vspace{0.75em}
\noindent\textbf{From projection-based reduced-order models (ROMs) to latent dynamics.}

This challenge has long motivated reduced-order modeling (ROM), which seeks to approximate
the full-order dynamics in a low-dimensional representation.
Projection-based approaches such as Proper Orthogonal Decomposition (POD) combined with
Galerkin or Petrov--Galerkin projection have proven effective for moderately nonlinear systems
near a fixed operating point \citep{berkooz1993proper, rowley2017model, benner2015survey}.
However, linear-subspace ROMs often struggle in the presence of strong nonlinearities,
transient dynamics, bifurcations, or wide parameter variations, where a single global basis
fails to capture the evolving solution manifold.
Recent advances in parametric and geometry-aware POD frameworks, including Grassmann-
and Stiefel-manifold formulations, have extended the applicability of linear subspace methods
by accounting for smooth subspace variation across parameters
\citep{padovan2024data, Sato_Schmidt_2025}.
Nevertheless, these approaches remain fundamentally tied to locally optimal linear subspaces
and typically require explicit interpolation or re-projection when traversing time,
parameters, or distinct dynamical regimes.

Recent advances in machine learning have given rise to a new class of ROMs in which linear
subspaces are replaced by nonlinear latent representations learned directly from data using
autoencoders or variational autoencoders
\citep{hinton2006reducing, kingma2013auto, brunton2020machine, fresca2021comprehensive}.
These methods learn compact latent coordinates that capture nonlinear structure in
high-dimensional solution manifolds, often achieving substantial compression while
maintaining high reconstruction fidelity.
In this setting, the central challenge shifts from representation learning to accurately
modeling the evolution of latent states.

Within latent-space ROMs, existing approaches to learning latent dynamics fall into two broad classes. The first consists of discrete-time, recurrent or autoregressive models such as AE--LSTM and AE--GRU that advance the latent state through small time increments learned from snapshot sequences \citep{maulik2021reduced,vlachas2018data,pathak2018model}. These models show strong short-term predictive capability for fluid and climate systems but rely on iterative rollout, so errors accumulate over long horizons and stability depends on the temporal resolution encoded in the training data \citep{vlachas2020backpropagation}. The second class adopts a continuous-time formulation based on Neural Ordinary Differential Equations (NODEs), in which a neural network learns a latent vector field that is integrated by an ODE solver \citep{chen2018neural,rubanova2019latent}. This yields time-continuous latent dynamics that are, in principle, independent of the training time step. \cite{farenga2025latent} formalize this perspective in a Latent Dynamics Model (LDM) framework and provide error and stability estimates for AE--NODE ROMs in multi-query settings, but practical deployment still requires recursive integration for long-horizon prediction, reintroducing step-size sensitivity and preventing direct single-shot mappings between distant time instances.

\vspace{0.75em}
\noindent\textbf{Operator learning in state and latent space.}

In parallel, neural operator learning seeks to approximate mappings between infinite-dimensional
function spaces.
Representative architectures such as Deep Operator Networks (DeepONets)
\citep{lu2019deeponet, lu2021learning} and Fourier Neural Operators (FNOs)
\citep{li2020fourier, anandkumar2020neural}
take an initial or boundary condition together with physical parameters as input and return
the solution evaluated at one or more target times or spatial locations.
Treating time as an explicit input enables single-shot evaluation
at arbitrary times without recursive temporal integration, and these operators have shown strong
performance across diverse parametric PDEs.

To improve efficiency and scalability, several works have combined operator learning with
nonlinear dimensionality reduction.
In these latent-operator approaches, high-dimensional solution fields are first compressed
using an autoencoder, an operator is learned in the latent space, and the output is decoded
back to the physical domain.
Latent DeepONet follows a two-stage design with a fixed autoencoder and a DeepONet acting on frozen latent codes \citep{kontolati2024learning}.
The Latent Neural PDE Solver couples a convolutional autoencoder with a latent propagator realized through spectral
and attention-based kernels \citep{li2025latent}.
Physics-informed latent operators further train encoder, operator, and decoder jointly using PDE residual losses to reduce data demands and promote physical consistency \citep{li2024physics, karumuri2026physics}.
Collectively, these approaches combine the expressive compression of autoencoders with
the single-shot evaluation capability of neural operators.

Despite their success, most operator-learning formulations are trained and evaluated
in an initial-value setting, where the operator maps from an initial condition
(or early-time state) and parameters to the solution at later times.
Although these architectures are, in principle, function-to-function maps that could
accept arbitrary states as input, their training distributions and benchmark problems
remain predominantly centered on initial-condition ensembles
\citep{li2020fourier, lu2021learning, li2025latent}.
As a result, querying such operators from intermediate observed states
along a trajectory can induce a nontrivial distribution shift.

This re-anchoring capability is typically not explicitly enforced during training,
even though it is essential for applications such as data assimilation,
closed-loop control, and model-predictive forecasting.
In particular, neural operators are commonly trained to learn mappings of the form
$ \boldsymbol{u}_0 \mapsto \boldsymbol{u}(t) $.
In contrast, the proposed \emph{Latent Evolution Operator} (LEO) framework learns a family of finite-time evolution operators
$ \mathfrak{F}_\tau: \boldsymbol{u}(t) \mapsto \boldsymbol{u}(t+\tau) $,
enabling prediction from arbitrary intermediate states without reliance on initial-condition anchoring.

\vspace{0.75em}
\noindent\textbf{Bridging the gap: a parametric latent evolution operator.}
These approaches leave a gap between continuous-time latent dynamics models and operator-learning methods. LDM-type approaches provide a principled framework for time-continuous latent evolution but require numerical integration and step-by-step rollout. Latent operator approaches enable single-shot evaluation across time horizons but are typically trained on a restricted class of inputs, most often initial conditions, rather than arbitrary anchor states along a trajectory. Neither paradigm alone fully addresses the need for efficient, anchor-agnostic state-to-state prediction in multi-query and closed-loop settings.

The goal of this work is to unify continuous-time latent dynamics and operator learning
by introducing the LEO, a latent-space approximation
of the finite-time evolution operator.
LEO defines a parameterized mapping that, given a system state at an arbitrary anchor time,
together with physical parameters and a forecast horizon, returns the future state
in a single evaluation while operating entirely within a learned low-dimensional latent manifold.
Conceptually, LEO acts as a global solution operator in latent space, directly mapping
between states separated by finite time intervals rather than approximating a local
vector field that must be recursively integrated.

\vspace{0.75em}
\noindent\textbf{Contributions.}
This work makes three contributions.

First, on the theoretical side, we formalize the notion of a LEO for parametric, time-dependent PDEs and establish an encoder--propagator--decoder factorization of the finite-time solution operator under mild manifold and regularity assumptions. We further show that this factorization can be approximated by ReLU networks with complexity governed by the intrinsic manifold dimension rather than the ambient state dimension.

Second, on the modeling side, we introduce an end-to-end trainable LEO architecture that learns a parameterized latent evolution operator mapping states across arbitrary time horizons. By training on state pairs sampled across trajectories, the framework enables anchor-agnostic, state-to-state prediction and supports variants with and without latent-consistency regularization to balance structural consistency and forecasting accuracy.

Third, on the empirical side, we evaluate LEO on three benchmark systems: the viscous Burgers' equation, a two-dimensional advection--diffusion system, and a stochastically forced shallow-water model. Across these cases, LEO achieves accurate single-shot forecasts, improved robustness to intermediate-state initialization, and substantial speedups over recurrent latent models, demonstrating its effectiveness for multi-query prediction tasks.

This formulation enables efficient and stable prediction in multi-query settings where repeated simulation or recursive rollout is computationally prohibitive.

We provide an open-source implementation at \textbf{GitHub}\footnote{\url{https://github.com/Khalid-Rafiq-01/Latent-Evolution-Operator-LEO-}} and an interactive demo at \textbf{Hugging Face Spaces}\footnote{\url{https://huggingface.co/spaces/krafiq/LEO}}.

\section{Methodology}
\label{sec:method}

\subsection{Problem setup}

We consider time-dependent parametric partial differential equations (PDEs) of the form
\begin{equation}
  \frac{\partial \boldsymbol{u}(\boldsymbol{x},t,\bzeta)}{\partial t}
  = \boldsymbol{F}\big(\boldsymbol{u}(\boldsymbol{x},t,\bzeta), t, \bzeta\big),
  \qquad \boldsymbol{x} \subset \mathbb{R}^d,\; t \in [0,T],
  \label{eq:param_pde_method}
\end{equation}
with initial condition
\(
\boldsymbol{u}(\boldsymbol{x},0,\bzeta) = \boldsymbol{u}_0(\boldsymbol{x},\bzeta).
\)
Here $\boldsymbol{F}$ denotes the PDE right-hand side, $\boldsymbol{u}(\boldsymbol{x},t,\bzeta)$ denotes the (possibly vector-valued) solution field of the PDE at spatial location $\boldsymbol{x}$ and time $t$, and $\bzeta$ denotes a vector of physical or nondimensional parameters (e.g.\ boundary conditions, forcing amplitudes, Reynolds number). After spatial discretization on a $n$-point grid, the solution at time $t$ is represented by a high-dimensional state vector
$\boldsymbol{u}(t,\bzeta) \in \mathbb{R}^n$, which we refer to as the discretized state.

For fixed $\bzeta$, the PDE induces a family of time--$\tau$ evolution operators
\[
\mathfrak{F}_\tau: \bu(t,\bzeta) \mapsto \bu(t+\tau,\bzeta),
\qquad \tau \ge 0,
\]
which together define the solution operator governing the system dynamics. Here, $\mathfrak{F}_\tau$ denotes the induced time--$\tau$ evolution operator
on the discretized state.

Our objective is to learn a data-driven surrogate for this family of operators that enables state prediction across a continuum of forecast horizons $\tau$ and parameter values $\bzeta$. In contrast to standard neural operator or sequence-learning approaches that map initial conditions to a fixed final time, we target a state--to--state, variable-horizon formulation. This allows forecasts from arbitrary intermediate states using only snapshot pairs, without requiring initial-condition trajectories or sequential rollout.

\subsection{Encoder--propagator--decoder factorization of the evolution operator}
In many parametric PDE systems, the solution set $\bu(t,\bzeta)$ concentrates near a low-dimensional manifold embedded in the ambient space $\RR^n$~\citep{cohen2015approximation}. This structure arises from physical constraints, smooth parametric dependence, and dissipative dynamics, which motivates us to represent the evolution in a reduced coordinate system. 

We assume that the collection of all PDE solutions lies on an $m$--dimensional compact Riemannian manifold $\cM$ that is isometrically embedded in $\RR^n$, and that the evolution operator $\frakF$ is Lipschitz continuous. The necessary
 manifold definitions are recalled in Section~\ref{appsub:pre}.

\begin{assumption}
\label{assumption:manifold}
Let $B>0$ and  assume that the collection of all PDE solutions in \eqref{eq:param_pde_method} lies on  an $m$-dimensional compact smooth Riemannian manifold $\cM\subset [-B,B]^n$  isometrically embedded in $\mathbb{R}^n$ with reach $\tau_{\cM}>0$.  Suppose the discretized PDE solution of \eqref{eq:param_pde_method} is driven by a Lipschitz\footnote{The evolutionary operator $\frakF$ is Lipschitz in the sense that $\|\frakF(\bu_1(t,\bzeta),\bzeta)- \frakF(\bu_2(t,\bzeta),\bzeta)\|\le \Lip_{\frakF}\|\bu_1(t,\bzeta)-\bu_2(t,\bzeta)\|$ for all $\bu_1(t,\bzeta),\bu_2(t,\bzeta)\in \cM$ and all $t,\tau,\bzeta$. Here $\Lip_{\frakF}$ denotes the Lipschitz constant of $\frakF$.} evolution operator 
 $$\bu(t+\tau,\bzeta) = \frakF(\bu(t,\bzeta),\bzeta)$$
 and all discretized solutions
 lie on $\cM$ such that $\{\bu( t, {\bzeta})\} \subset \cM $. 
\end{assumption}

Under Assumption~\ref{assumption:manifold}, there exists an \emph{oracle latent factorization} consisting of an encoder--decoder pair $(\cE^\star,\cD^\star)$ that parameterizes $\cM$, together with a latent propagator $\cP^\star$ that governs the PDE evolution in the latent space.

 \begin{theorem}
 \label{thm:decom}
Under Assumption \ref{assumption:manifold}, there exist a Lipschitz\footnote{We denote the Lipschitz constants of $\cE^*$ and $\cD^*$ by $\Lip_{\cE^*}$ and $\Lip_{\cD^*}$ such that $\|\cE^*(\bu_1)-\cE^*(\bu_2)\|\le \Lip_{\cE^*}\|\bu_1-\bu_2\|$ for $\bu_1,\bu_2\in\cM$ and $\|\cD^*(\bz_1)-\cD^*(\bz_2)\|\le \Lip_{\cD^*}\|\bz_1-\bz_2\|$ for $\bz_1,\bz_2 \in {\rm Range}(\cE^*(\cM))$.} 
oracle encoder and decoder pair $\cE^*:\cM\rightarrow \RR^{C_{\cM}(m+1)}$ and $\cD^*: {\rm Range}(\cE^*(\cM)) \rightarrow \cM$ satisfying $\cD^*\circ\cE^*(\bu)=\bu, \ \forall \bu\in \cM,$
     where $C_{\cM}$ denotes the number of charts for $\cM$. Furthermore, there exists a Lipschitz  latent propagator $\cP^*$ such that $\bz(t,\bzeta) = \cE^*(\bu(t,\bzeta))$ and
         \begin{align}
\bz(t+\tau,\bzeta) &= 
 \cP^*(\bz(t,\bzeta),\tau,\bzeta) = \cE^*\circ\frakF(\cD^*(\bz(t,\bzeta)),\bzeta)
 \label{eq:oracleprop}
 \\
 \bu(t+\tau,\bzeta) &= \frakF(\bu(t,\bzeta),\bzeta)=\cD^*\circ\cP^*(\cE^*(\bu(t,\bzeta)),\tau,\bzeta)
  \label{eq:oracleevolve}
         \end{align}
         for all $t,\tau,\bzeta$.
 \end{theorem}

\begin{wrapfigure}{r}{0.42\textwidth}
\vspace{-1.0em}
\centering
\includegraphics[width=\linewidth]{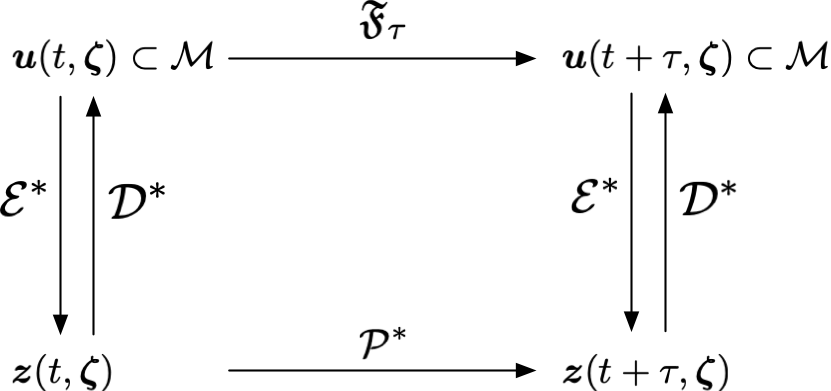}
\caption{The PDE evolution operator $\frakF$ and its encoder--propagator--decoder factorization.}
\label{fig:workflow}
\vspace{-1.0em}
\end{wrapfigure}

Theorem \ref{thm:decom} is proved in Appendix \ref{appsub:proofthmdecom}. Theorem \ref{thm:decom} shows that, under a manifold hypothesis of the solution manifold $\cM$, the Lipschitz evolutionary operator $\frakF$ exhibits an encoder-propagator-decoder decomposition in \eqref{eq:oracleevolve}. The encoder $\cE^*$ and decoder $\cD^*$ transforms the discretized PDE solution to a latent state of dimension $C_{\cM}(m+1)$ where $m$ is the intrinsic dimension of the solution manifold $\cM$, and $C_{\cM}$ characterizes the complexity of the manifold $\cM$.  Then the PDE evolution is driven by the latent propagator $\cP^*$.  A workflow about this decomposition is illustrated in Figure  \ref{fig:workflow}.

 \subsection{Latent evolution--operator network representation}

The encoder $\cE^*$, decoder $\cD^*$ and propagator $\cP^*$ are called the oracle encoder, decoder and propagator respectively, which depend on the underlying manifold $\cM$ and the groundtruth evolution operator $\frakF$. In this paper, we represent these oracle operators by neural networks $(\cE_{\theta_e},\cD_{\theta_d},\cP_{\theta_p})$.

 We consider feedforward neural networks (FNNs) with a nonlinear activiation function $\sigma$. An FNN with $L$ layers is defined as
  
\begin{equation}
 f(\bu)= W_L\cdot{\sigma}\left( W_{L-1}\cdots {\sigma}(W_1\bu+\bb_1)+ \cdots +\bb_{L-1}\right)+\bb_L,
 \label{eq.ReLU}
\end{equation}
   
 where $W_i$ and $\bb_i$ denote the weight matrices and bias vectors, respectively, and $\sigma$ is applied element-wise. When $\sigma(x) = \max(x, 0)$, the FNN is referred to as a ReLU network.

 We define a class of neural networks with inputs in $\mathbb{R}^{d_1}$ and outputs in $\mathbb{R}^{d_2}$ as
 
\begin{equation}
\begin{aligned}
\cF_{\mathrm{NN}}(d_1,d_2;L,p,K,\kappa,R)
= \Bigl\{ f:\RR^{d_1}\rightarrow \RR^{d_2}\ :\;&
f \text{ has the form of } \eqref{eq.ReLU}, \\
& f \text{ has } L \text{ layers and width bounded by } p, \\
& \|f\|_{L^{\infty}} \leq R, \\
& \sum_{i=1}^L \bigl( \|W_i\|_0 + \|\bb_i\|_0 \bigr) \leq K, \\
& \|W_i\|_{\infty,\infty} \leq \kappa,\quad
\|\bb_i\|_{\infty} \leq \kappa,\quad
i=1,\dots,L
\Bigr\}.
\end{aligned}
\end{equation}

 where $\|{H}\|_{\infty, \infty} = \max_{i, j} |H_{ij}|$ for a matrix $H$, $\|\cdot\|_0$ denotes the number of non-zero elements of its argument, and $\|f\|_{L^\infty}$ denotes the ${L^\infty}$ norm of the function $f$ on the domain.

 Our main theoretical result in Theorem~\ref{thm:nn} demonstrates the
 representation capability of neural networks for the evolution of the PDE in Eq.~\eqref{eq:parametric_pde} under the manifold hypothesis.

 \begin{theorem}
 \label{thm:nn}
 Under Assumption \ref{assumption:manifold}, for any $\varepsilon\in (0,1)$, there exists 
 an encoder ReLU \footnote{The activation function is the rectified linear unit $\sigma(a)=\max\{a,0\}$.} network $\cE_{\theta_e}\in \cF_{\rm NN}(n,{C_{\cM}}(m+1),L_{E},p_{E},K_{E},\kappa_{E},M_E)$,
 a decoder ReLU network
 $\cD_{\theta_d} \in \cF_{\rm NN}(C_{\cM}(m+1),n,L_{D},p_{D},K_{D},\kappa_{D},M_D)$, and a propagator ReLU network $\cP_{\theta_p}\in \cF_{\rm NN}(C_{\cM}(m+1),C_{\cM}(m+1),L_P,p_P,K_P,\kappa_P,M_P)$
 with 
 \begin{equation}
 \begin{aligned}
             &L_{E}=O(\log \varepsilon^{-1}), \ p_{E}=O(\varepsilon^{-m}), \\ 
             & \ K_{E}=O(\varepsilon^{-m}\log \varepsilon^{-1}), \ \kappa_{E}= O(\varepsilon^{-m}), \ M_{E}=O(1)
             \\
             &L_{D}=O(\log \varepsilon^{-1}), \ p_{D}=O(\varepsilon^{-m}), \\ 
             & \ K_{D}=O(\varepsilon^{-m}\log \varepsilon^{-1}), \ \kappa_{D}= O(\varepsilon^{-m}), \ M_{D}=B
             \\
              &L_P=O(\log \varepsilon^{-1}), \ p_P=O(\varepsilon^{-m}), \\ 
              & \ K_P=O(\varepsilon^{-m}\log \varepsilon^{-1}), \ \kappa_P= O(\varepsilon^{-m})
             \ M_P= O(1)
         \end{aligned}
         \end{equation}
         such that
            \begin{equation}
       \begin{aligned}
 \left\|\cD_{\theta_d}\circ\cP_{\theta_p}(\cE_{\theta_e}(\bu(t,\bzeta)),\tau,\bzeta) - \frakF(\bu(t,\bzeta),\bzeta)\right\| 
 \\
   \le
  \left(\Lip_{\cD_{\theta_d}} \Lip_{\cP_{\theta_p}}
  % term 2
  +\Lip_{\cD_{\theta_d}}
  % term 3
  +1\right)\varepsilon.
  \label{eq:errorthm:nn}
        \end{aligned}
         \end{equation}
 The constants hidden in $O(\cdot)$ depend on $m,n,\tau_{\cM}$, $B$, $\Lip_{\frakF}$, $C_{\cM}$ (the number of charts for $\cM$) and the output, derivatives of the coordinate maps $\phi_i$'s of $\cM$ and the partition of unity $\rho_i$'s up to order $1$. In Eq.~\eqref{eq:errorthm:nn}, $\Lip_{\cD_{\theta_d}}$  and $\Lip_{\cP_{\theta_p}}$ denote the Lipschitz constants\footnote{We consider at atlas of $\cM$, denoted by $\{(U_i,\phi_i)\}_{i=1}^{C_{\cM}}$ such that each $\phi_i$ is an orthogonal projection onto a local tangent space, as constructed in the proof of Theorem \ref{thm:nn} in Appendix \ref{appsub:proofthmnn}. Denote $b =\max\left(\sup_{\bu(t,\bzeta)\in\cM }\max_j \|\phi_j(\bu(t,\bzeta))\|_{\infty},1\right).$ The Lipschitz constant of $\cD_{\theta_d}$ is defined such that $\|\cD_{\theta_d}(\bz_1) - \cD_{\theta_d}(\bz_2)\| \le \Lip_{\cD_{\theta_d}} \|\bz_1-\bz_2\|$ for all $\bz_1,\bz_2\in [-b,b]^{C_{\cM}(m+1)} \subset \RR^{C_{\cM}(m+1)}$.
 The Lipschitz constant of $\cP_{\theta_p}$ is defined such that $\|\cP_{\theta_p}(\bz_1) - \cP_{\theta_p}(\bz_2)\| \le \Lip_{\cP_{\theta_p}} \|\bz_1-\bz_2\|$ for all $\bz_1,\bz_2\in [-b,b]^{C_{\cM}(m+1)}$.} of the decoder network and the propagator network respectively.
 \end{theorem}

Theorem \ref{thm:nn} is proved in Appendix \ref{appsub:proofthmnn}. 
 Theorem \ref{thm:nn} establishes the representational capacity of neural networks within the encoder-propagator-decoder framework to approximate the evolutionary operator $\mathfrak{F}_\tau$. As the approximation accuracy $\varepsilon$ varies, the network size in Theorem \ref{thm:nn} scales in a power law of $\varepsilon$, and importantly, the power depends on the intrinsic dimension $m$ instead of the ambient dimension $n$. 
 
\begin{figure}[t]
   \centering
   \includegraphics[width=\textwidth]{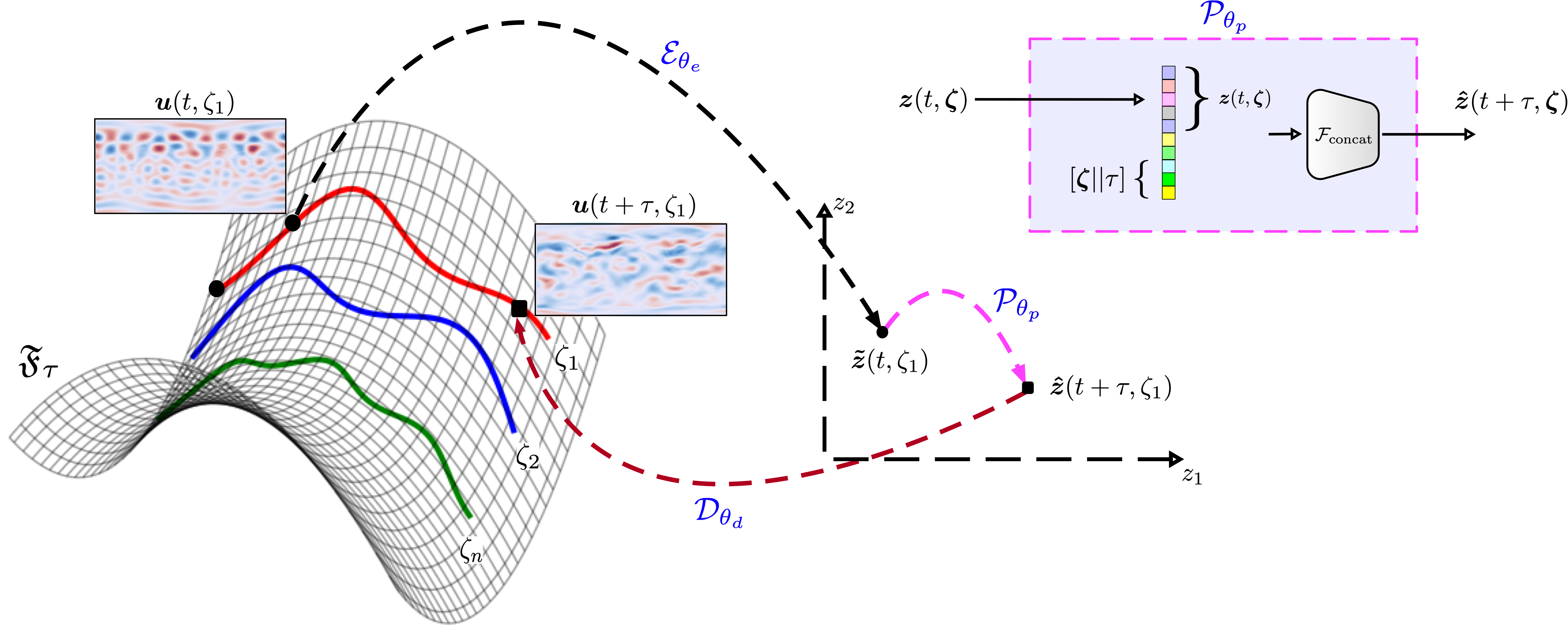}
   \caption{An illustration of the Latent Evolution Operator (LEO) framework. On the solution manifold
   $\mathcal{M} \subseteq \mathbb{R}^n$, three solution trajectories correspond to the parameters
   $\zeta_1,\zeta_2,...\zeta_n$. For a fixed parameter $\zeta_1$, the
   evolution operator advances the state along the red trajectory from $\bu(t,\zeta_1)$ to $\bu(t+\tau,\zeta_1)$. The encoder $\mathcal{E}_{\theta_e}$ maps $\bu(t,\zeta_1)$
   to a latent state $\tilde \bz(t,\zeta_1)$ in the low--dimensional latent space. The latent propagator $\mathcal{P}_{\theta_p}$, implemented by the concatenation network
   $\mathcal{F}_{\mathrm{concat}}$ inside the shaded block, takes the latent state $\tilde \bz(t,\zeta_1)$ together with the parameter vector and forecast horizon
   $[\bzeta_1 \| \tau]$ and produces the propagated
   (parameter--informed) latent state $\hat \bz(t +\tau,\zeta_1)$
   (magenta arrow). The  decoder $\mathcal{D}_{\theta_d}$ then maps
   $\hat \bz(t +\tau,\bzeta_1)$ back to the forecasted state
   $\hat \bu(t+\tau,\bzeta_1)$ (maroon dashed curve).
   }
   \label{fig:main_idea}
\end{figure}

With the encoder, decoder and propagator networks $(\cE_{\theta_e},\cD_{\theta_d},\cP_{\theta_p})$, the inference of the evolution operator $\frakF$ can be performed in the latent space. Given the PDE state $\bu(t,\bzeta)$, parameter $\bzeta$ and forecast horizon $\tau$, we can predict $\bu(t+\tau,\bzeta)$ by
$$\hat\bu(t+\tau,\bzeta) = \cD_{\theta_d}\circ\cP_{\theta_p}(\cE_{\theta_e}(\bu(t,\bzeta)),\tau,\bzeta),$$
illustrated by Figure \ref{fig:main_idea}. Here, \(\hat{\boldsymbol{u}}(t+\tau,\bzeta)\) denotes the forecasted state produced by the neural networks and is intended to approximate the true state \(\boldsymbol{u}(t+\tau,\bzeta)\). We use \(\tilde{\boldsymbol{z}}(t,\bzeta) = \cE_{\theta_e}(\boldsymbol{u}(t,\bzeta))\) for the latent representation of the state obtained from the learned encoder, to distinguish it from the oracle latent variable \(\boldsymbol{z}(t,\bzeta)\) in Theorem~\ref{thm:decom}.

\subsection{Data and Training}
\label{data_training}

We construct the training dataset from $K$ distinct parameter configurations
$\{\zeta_k\}_{k=1}^K$.
For each parameter $\zeta_k$, we select $J$ anchor times
$\{t_{kj}\}_{j=1}^J \subset [0,T]$ along the trajectory, and for each anchor time choose $I$ forecast horizons
$\{\tau_{kji}\}_{i=1}^I$. Anchor times \(t_{kj}\) will later be denoted generically by \(t_a\) when
we focus on individual trajectories.
Each simulation thus provides many state--to--state supervision pairs at different temporal offsets. The resulting dataset is
{\small
\begin{equation}
\mathcal{S}
  :=
  \Big\{
      \big(\boldsymbol{u}(t_{kj},\zeta_k),\;
           \boldsymbol{u}(t_{kj}+\tau_{kji},\zeta_k)\big)
      : i=1,\ldots,I,\; j=1,\ldots,J
  \Big\}_{k=1}^K.
  \label{eq:dataset}
\end{equation}
}
This construction enables combinatorial reuse of trajectories, generating a large set of state-to-state training pairs across multiple forecast horizons without requiring additional simulations. In practice, anchor times $t_{kj}$ are sampled uniformly along each trajectory,
and forecast horizons $\tau_{kji}$ are drawn from a prescribed range
$\tau \in [\tau_{\min},\tau_{\max}]$.
This sampling strategy ensures coverage of both short- and long-horizon dynamics;
additional details are provided in Appendix~\ref{appendix:data_split}.

Given the dataset in \eqref{eq:dataset}, we introduce a unified training objective that balances data-space accuracy, latent-space consistency, and prior regularization. Adjusting the relative weights of these terms yields a family of LEO that interpolate between strictly consistent latent representations and more flexible direct models. The unified objective is given as

\begin{equation}
\begin{aligned}
L(\bu,t,\tau,\bzeta)
&= L_{\mathrm{rec}}
 + \lambda_{\mathrm{dir}} L_{\mathrm{dir}}
 + \lambda_{\mathrm{con}} L_{\mathrm{con}}
 + \beta L_{\mathrm{KL}},
\end{aligned}
\label{eq:unified_loss_compact}
\end{equation}

{\small
\begin{equation}
\begin{aligned}
L_{\mathrm{rec}}
&= \frac{1}{KJ}\sum_{k=1}^K\sum_{j=1}^J
\big\|\bu(t_{kj},\zeta_k)
- \mathcal{D}_{\theta_d}\!\big(\mathcal{E}_{\theta_e}(\bu(t_{kj},\zeta_k))\big)
\big\|_2^2,
\end{aligned}
\end{equation}

\begin{equation}
\begin{aligned}
L_{\mathrm{dir}}
&= \frac{1}{KJI}\sum_{k=1}^K\sum_{j=1}^J\sum_{i=1}^I
\big\|\bu(t_{kj}+\tau_{kji},\zeta_k)
\\
&- \mathcal{D}_{\theta_d}\!\big(
\mathcal{P}_{\theta_p}(
\mathcal{E}_{\theta_e}(\bu(t_{kj},\zeta_k)),\tau_{kji},\zeta_k)
\big)\big\|_2^2,
\end{aligned}
\end{equation}
\begin{equation}
\begin{aligned}
L_{\mathrm{con}}
&= \frac{1}{KJI}\sum_{k=1}^K\sum_{j=1}^J\sum_{i=1}^I
\big\|
\mathcal{P}_{\theta_p}(
\mathcal{E}_{\theta_e}(\bu(t_{kj},\zeta_k)),\tau_{kji},\zeta_k) \\
& - \mathcal{E}_{\theta_e}(\bu(t_{kj}+\tau_{kji},\zeta_k))
\big\|_2^2,
\end{aligned}
\end{equation}
\begin{equation}
\begin{aligned}
L_{\mathrm{KL}}
&= \mathrm{KL}\!\Big(
q\big(\mathcal{E}_{\theta_e}(\bu(t,\bzeta))\big)
\,\big\|\, \mathcal{N}(0,I)\Big).
\label{eq:loss_components}
\end{aligned}
\end{equation}
}

where $q(\cdot)$ denotes the encoder--induced approximate posterior. While $L_{\mathrm{dir}}$ enforces prediction accuracy in solution space, $L_{\mathrm{con}}$ controls identifiability and alignment of the latent state, mitigating non-uniqueness in latent factorizations that can arise when supervision is imposed only in the data space.

The reconstruction loss $L_{\mathrm{rec}}$ trains the encoder--decoder pair to provide a faithful low--dimensional representation of the solution manifold by penalizing reconstruction error at the anchor time $t_a$.
The direct forecasting loss $L_{\mathrm{dir}}$ promotes accurate prediction in the data space by training the composed map
$\cD_{\theta_d}\circ\cP_{\theta_p}\circ\cE_{\theta_e}$ to approximate the nonlinear evolution operator
$\frakF$.
The latent--consistency loss $L_{\mathrm{con}}$ promotes consistency in the latent space between the propagated latent state
$\cP_{\theta_p}(\cE_{\theta_e}(\bu(t,\bzeta)),\tau,\bzeta)$
and the latent encoding of the true future state
$\cE_{\theta_e}(\bu(t+\tau,\bzeta))$, thereby encouraging a single approximately consistent latent trajectory aligned with the PDE flow, mirroring the oracle construction in Theorem~\ref{thm:nn}.
Finally, the KL term $L_{\mathrm{KL}}$ is the standard variational regularizer, encouraging the encoder-induced posterior to remain close to a unit Gaussian prior and promoting smooth, non-degenerate latent representations.

\begin{figure}[t]
  \centering
  \includegraphics[width=\textwidth]{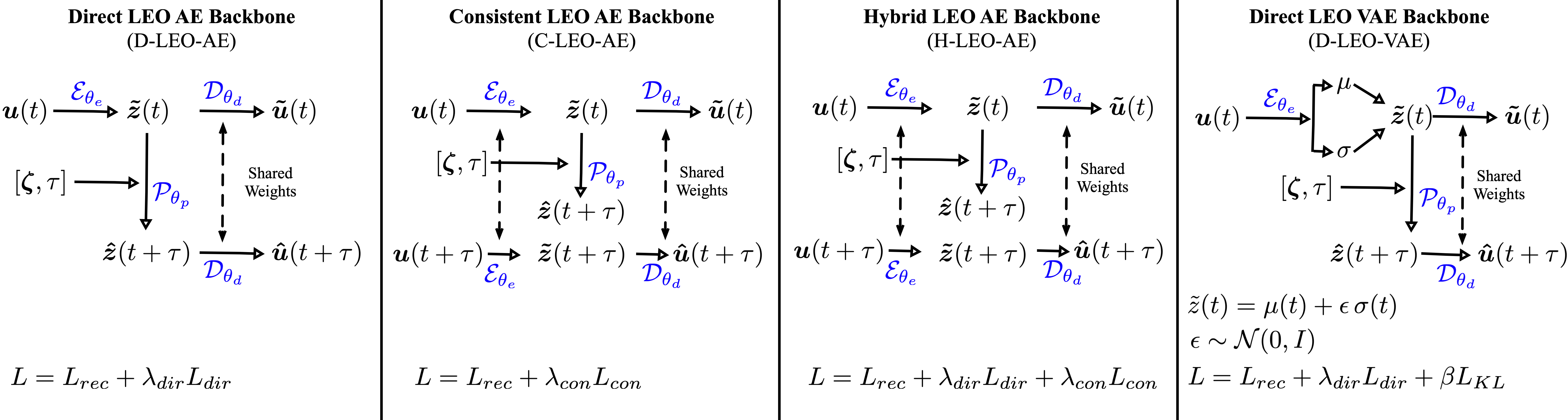}
  \caption{\textbf{Latent Evolution--Operator (LEO) variants.}
  Schematic comparison of the four architectures considered in this work.
  In all cases an encoder $\mathcal{E}_{\theta_e}$ maps the input state $\bu(t)$
  to a latent code $\tilde \bz(t)$. The propagator $\mathcal{P}_{\theta_p}$
  advances the latent conditioned on parameter inputs $[\bzeta,\tau]$ to
  $\hat \bz(t+\tau)$, and a decoder $\mathcal{D}_{\theta_d}$ reconstructs states.
  Decoder is shared for both reconstruction $\tilde \bu(t)$ and
  forecasts $\hat \bu(t+\tau)$.
  \emph{Direct LEO AE backbone (D--LEO--AE)} uses only reconstruction and
  direct forecasting with loss
  $L = L_{\mathrm{rec}} + \lambda_{\mathrm{dir}} L_{\mathrm{dir}}$.
  \emph{Consistent LEO AE backbone (C--LEO--AE)} adds a latent consistency
  term enforcing approximate commutativity
  $\mathcal{P}\circ\mathcal{E} \approx \mathcal{E}\circ \frakF$,
  with loss
  $L = L_{\mathrm{rec}} + \lambda_{\mathrm{con}} L_{\mathrm{con}}$.
  \emph{Hybrid LEO AE backbone (H--LEO--AE)} combines both supervision
  signals,
  $L = L_{\mathrm{rec}} + \lambda_{\mathrm{dir}} L_{\mathrm{dir}}
       + \lambda_{\mathrm{con}} L_{\mathrm{con}}$.
  \emph{Direct LEO VAE backbone (D--LEO--VAE)} replaces the deterministic
  encoder with a stochastic one,
  $\tilde \bz(t) = \mu(t) + \epsilon \sigma(t)$ with
  $\epsilon \sim \mathcal{N}(0,I)$,
  and augments the direct loss with a KL regularizer,
  $L = L_{\mathrm{rec}} + \lambda_{\mathrm{dir}} L_{\mathrm{dir}}
       + \beta L_{\mathrm{KL}}$,
  encouraging a well--structured latent space.
  }
  \label{fig:leo_variants}
\end{figure}

Different choices of the regularization parameters, $\lambda_{\mathrm{dir}}$, $\lambda_{\mathrm{con}}$ and $\beta$, recover the four LEO variants summarized in Fig.~\ref{fig:leo_variants}, ranging from the most constrained consistent formulation, which closely follows the oracle structure implied by Theorem~\ref{thm:nn}, to more flexible direct variants that relax latent consistency and/or the Gaussian prior. All variants share the same architectural backbone
$(\cE_{\theta_e},\cD_{\theta_d},\cP_{\theta_p})$ and differ only through their training objectives and supervision signals.

For the consistent and hybrid variants, we employ a staged training schedule in which the latent-consistency weight $\lambda_{\mathrm{con}}$ is held at zero during an initial warm-up phase and then linearly increased to its target value. This allows the autoencoder to first learn a stable latent representation before enforcing alignment between encoded and propagated states. The direct variants are trained with $\lambda_{\mathrm{con}}=0$ and a fixed $\lambda_{\mathrm{dir}}$. The encoded future
$\tilde \bz(t+\tau,\bzeta) = \cE_{\theta_e}(\bu(t+\tau,\bzeta))$ and the propagated latent
$\hat \bz(t+\tau,\bzeta)=\cP_{\theta_p}\big(\cE_{\theta_e}(\bu(t,\bzeta)),\tau,\bzeta\big)$
are constrained only through their decoded images, allowing functionally equivalent but misaligned latent coordinates. By contrast, enforcing $L_{\mathrm{con}}$ penalizes the mismatch, i.e., 
$\hat \bz(t+\tau,\bzeta)-\tilde \bz(t+\tau,\bzeta)$, encouraging latent dynamics that satisfy
\[
\cE_{\theta_e}\big(\frakF(\bu(t,\bzeta),\bzeta)\big)
\;\approx\;
\cP_{\theta_p}\big(\cE_{\theta_e}(\bu(t,\bzeta)),\tau,\bzeta\big).
\]
In Section~\ref{sec:latent_consistency}, we quantify this effect by comparing the separation between encoded and propagated latents, showing that consistent and hybrid LEO variants yield tightly aligned charts, whereas direct variants realize the same physical fields through more widely separated, yet functionally equivalent, latent representations.

\subsection{Train--validation splits and zones}
\label{sec:train_val_zones}

% We construct train--validation splits in the joint \((\bzeta,\tau)\) space. Given a collection of parameter--horizon pairs \((\bzeta,\tau)\), we first
% partition them into disjoint training and validation sets using a
% 70--30 split over \(\bzeta\). The validation set is then stratified into
% three regimes along each axis: \emph{left extrapolation} (LE),
% \emph{interpolation} (IN), and \emph{right extrapolation} (RE).

% \begin{figure}[h]
%     \begin{center}    
%     \includegraphics[width=0.95\columnwidth]{Fig4.png}
%   \end{center}
%     \caption{Train--validation partitioning in a representative
%     \((\zeta_1,\tau)\) plane. Hatched cells denote training pairs; gray
%     cells denote held-out validation pairs, stratified into nine zones
%     (LE/IN/RE along each axis).}
%     \label{fig:data_split_Burgers'}
% \end{figure}
% Along the parameter axis, LE contains values of \(\bzeta\) below the
% training range, IN contains values inside the training range, and RE
% contains values above it. Along the temporal axis, LE, IN, and RE are
% defined analogously with respect to the forecast horizon \(\tau\).
% Combining these regimes yields a \(3\times3\) grid of nine joint zones in
% a representative \((\zeta_1,\tau)\) plane, where \(\zeta_1\) denotes a
% scalar component of \(\bzeta\).

We construct train--validation splits in the joint \((\bzeta,\tau)\) space. Given a collection of parameter--horizon pairs \((\bzeta,\tau)\), we first
partition them into disjoint training and validation sets using a
70--30 split over \(\bzeta\). The validation set is then stratified into
three regimes along each axis: \emph{left extrapolation} (LE),
\emph{interpolation} (IN), and \emph{right extrapolation} (RE).

\begin{wrapfigure}{r}{0.50\textwidth}
    \vspace{-0.8em}
    \centering
    \includegraphics[width=\linewidth]{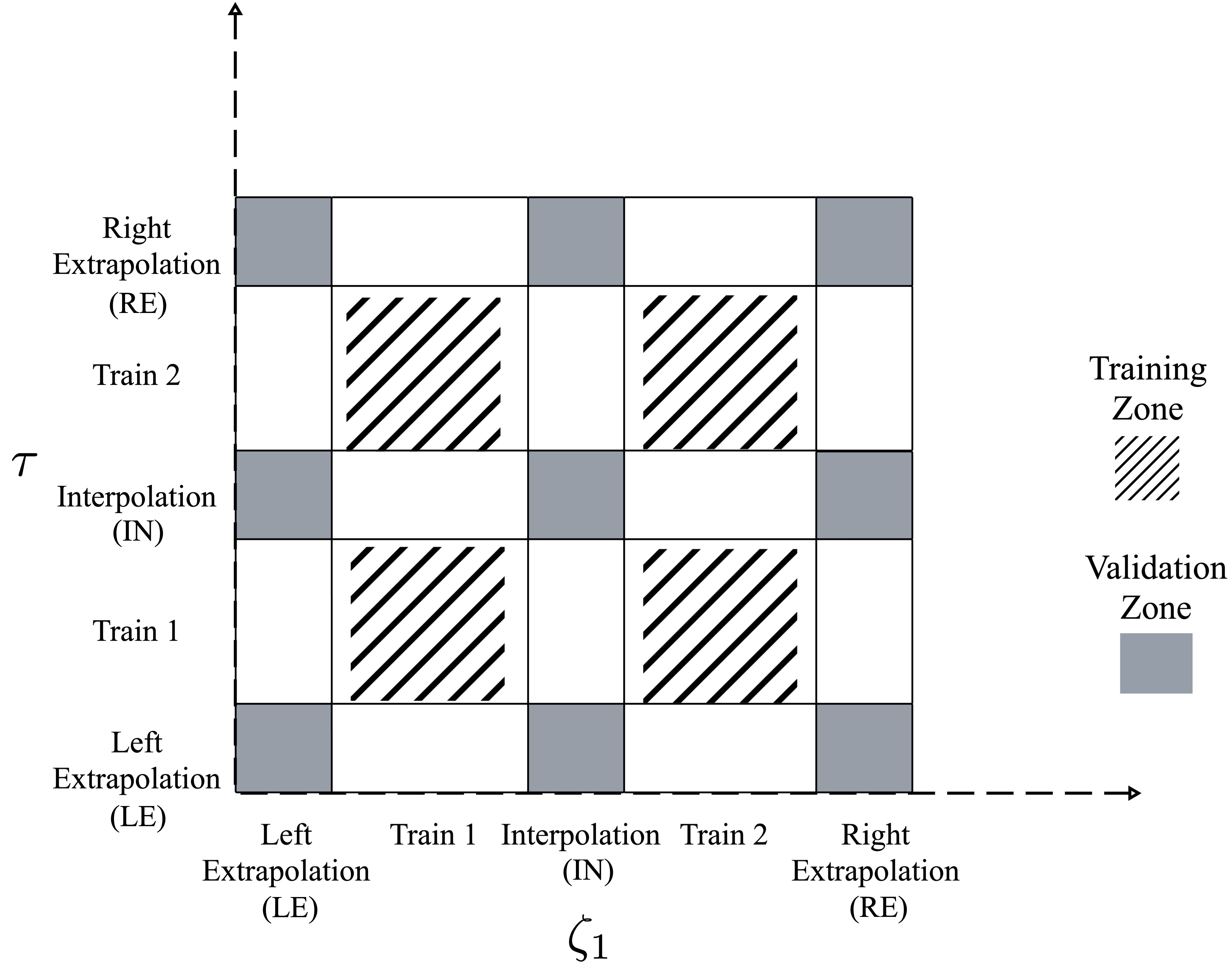}
    \caption{Train--validation partitioning in a representative
    \((\zeta_1,\tau)\) plane. Hatched cells denote training pairs; gray
    cells denote held-out validation pairs, stratified into nine zones
    (LE/IN/RE along each axis).}
    \label{fig:data_split_Burgers'}
    \vspace{-1.0em}
\end{wrapfigure}

Along the parameter axis, LE contains values of \(\bzeta\) below the
training range, IN contains values inside the training range, and RE
contains values above it. Along the temporal axis, LE, IN, and RE are
defined analogously with respect to the forecast horizon \(\tau\).
Combining these regimes yields a \(3\times3\) grid of nine joint zones in
a representative \((\zeta_1,\tau)\) plane, where \(\zeta_1\) denotes a
scalar component of \(\bzeta\).

We train LEO only on pairs drawn from the training band (hatched cells in
Fig.~\ref{fig:data_split_Burgers'}) and report validation errors
separately in each of the nine zones. This design separates
in-distribution generalisation (IN--IN) from various extrapolation
regimes, mirroring deployment settings where only a subset of operating
conditions can be simulated in advance.

Table~\ref{tab:notation} summarizes the main notation used throughout the paper.

\begin{table}[t]
\centering
\footnotesize
\setlength{\tabcolsep}{6pt}
\renewcommand{\arraystretch}{1.15}
\begin{tabularx}{\columnwidth}{@{} l l X @{}}
\toprule
 & \textbf{Symbol} & \textbf{Meaning} \\
\midrule

\multirow{6}{*}{\makecell[l]{PDE \\ \& evolution}}
& $\bu(\bx, t, \bzeta)$ & State vector at space $\bx$, time $t$, parameter $\bzeta$. \\
& $\bu(t, \bzeta)\in\mathbb{R}^n$ & Spatially discretized state vector at time $t$. \\
& $\boldsymbol{F}(\cdot)$ & PDE right-hand side.\\
& $\mathfrak{F}_\tau(\bu(t, \bzeta),\bzeta)$ & Time-$\tau$ evolution: $\bu(t+\tau, \bzeta)$. \\
& $\bzeta$ & Parameter vector (e.g., Reynolds number). \\
& $\tau$ & Forecast horizon (time offset). \\

\midrule
\multirow{6}{*}{\makecell[l]{LEO \\ components}}
& $\mathcal{E}_{\theta_e}$ & Encoder mapping $\bu(t, \bzeta)\mapsto \tilde \bz(t, \bzeta)$. \\
& $\mathcal{P}_{\theta_p}$ & Latent propagator mapping $\tilde \bz(t, \bzeta) \mapsto \hat \bz(t+\tau, \bzeta)$. \\
& $\mathcal{D}_{\theta_d}$ & Decoder mapping latent to state: $\tilde \bz(t, \bzeta) \mapsto \tilde \bu(t, \bzeta)$. \\
& $\tilde \bz(t, \bzeta)$ & Encoded latent: $\tilde \bz(t, \bzeta)=\mathcal{E}_{\theta_e}(\bu(t, \bzeta))$. \\
& $\hat \bz(t+\tau, \bzeta)$ & Propagated latent: $\hat \bz(t+\tau, \bzeta)=\mathcal{P}_{\theta_p}(\tilde \bz(t, \bzeta),\tau,\bzeta)$. \\
& $\hat \bu(t+\tau,\bzeta)$ & Forecasted state: $\hat \bu(t+\tau, \bzeta)=\mathcal{D}_{\theta_d}(\hat \bz(t+\tau, \bzeta))$. \\

\midrule
\multirow{3}{*}{\makecell[l]{Times \\ \& states}}
& $t_0$ & Initial-condition time ($t_0=0$). \\
& $t_a$ & Anchor time. \\
& $\bu(\bx, t_a)$ & Anchor state: $\bu(\bx, t_a)=\bu(\bx,t_a, \bzeta)$. \\
\bottomrule
\end{tabularx}
\caption{Summary of notation used in the LEO formulation and experiments.}
\label{tab:notation}
\end{table}

In the next section, we apply LEO to a canonical nonlinear benchmark,
the 1D viscous Burgers' equation, to assess forecasting accuracy,
latent consistency, and data scaling under controlled variation in Reynolds number $\mathrm{Re}$ and forecast time $\tau$. We also compare LEO against latent neural--operator baselines and examine how the learned latent
coordinates relate to the underlying transport and diffusion dynamics.

\section{Experiments on 1D viscous Burgers' equation}
\label{main_results_Burgers'}

The 1D viscous Burgers' equation serves as a canonical benchmark for nonlinear wave propagation and shock formation under varying viscosity \citep{maulik2021reduced}. The governing equation is
\begin{equation}
\frac{\partial u}{\partial t} + u \frac{\partial u}{\partial x}
= \frac{1}{\mathrm{Re}} \frac{\partial^2 u}{\partial x^2},
\end{equation}
subject to the initial and boundary conditions
\[
u(x, 0) = u_0(x), \quad u(0, t) = u(L, t) = 0, \quad x \in [0, L].
\]
Here $u(x,t)$ denotes the scalar velocity field and $\mathrm{Re}$ is the Reynolds number.

The spatial domain $[0,L]$ with $L=1$ is discretized using 128 uniform grid points, yielding a high-dimensional state vector $\mathbf{u}(t) \in \mathbb{R}^{128}$ at each time step, with a fixed time step $\Delta t = 0.004$.

We consider a smooth initial condition for which a closed-form solution exists via the Cole--Hopf transformation. The resulting analytical solution is
\begin{equation}
u(x, t) = \frac{x}{t + 1 + \sqrt{\frac{t + 1}{t_*}}
\exp\!\left(\frac{\mathrm{Re}\, x^2}{4(t + 1)}\right)}, \quad
\text{where } t_* = \exp\!\left(\frac{\mathrm{Re}}{8}\right).
\end{equation}
This exact solution provides a reference for evaluating both reconstruction and forecasting accuracy. The nonlinear dynamics of Burgers' flow, particularly at high $\mathrm{Re}$, generate sharp gradients that pose a challenge for data-driven forecasting models \citep{maulik2021reduced}.

We consider $\mathrm{Re} \in [400, 2400]$ and forecast horizons $\tau \in [0.6, 1.8]$, corresponding to approximately $30\%$--$90\%$ of the total wave propagation time across the domain. Lower $\mathrm{Re}$ values yield smooth, diffusion-dominated dynamics, whereas higher $\mathrm{Re}$ values produce sharp, advection-dominated profiles.

A detailed analysis of the latent dimension selection (with $m=2$ for Burgers') and the hyperparameter search is provided in Appendix~\ref{appendix:latent_dimension_choice}.

\subsection{LEO variant comparison}
% Ablation study on our LEO Family!
\subsubsection{Direct forecasting performance}
\label{sec:direct_forecasting}

\begin{figure}[t]
  \centering
  \includegraphics[width=\textwidth]{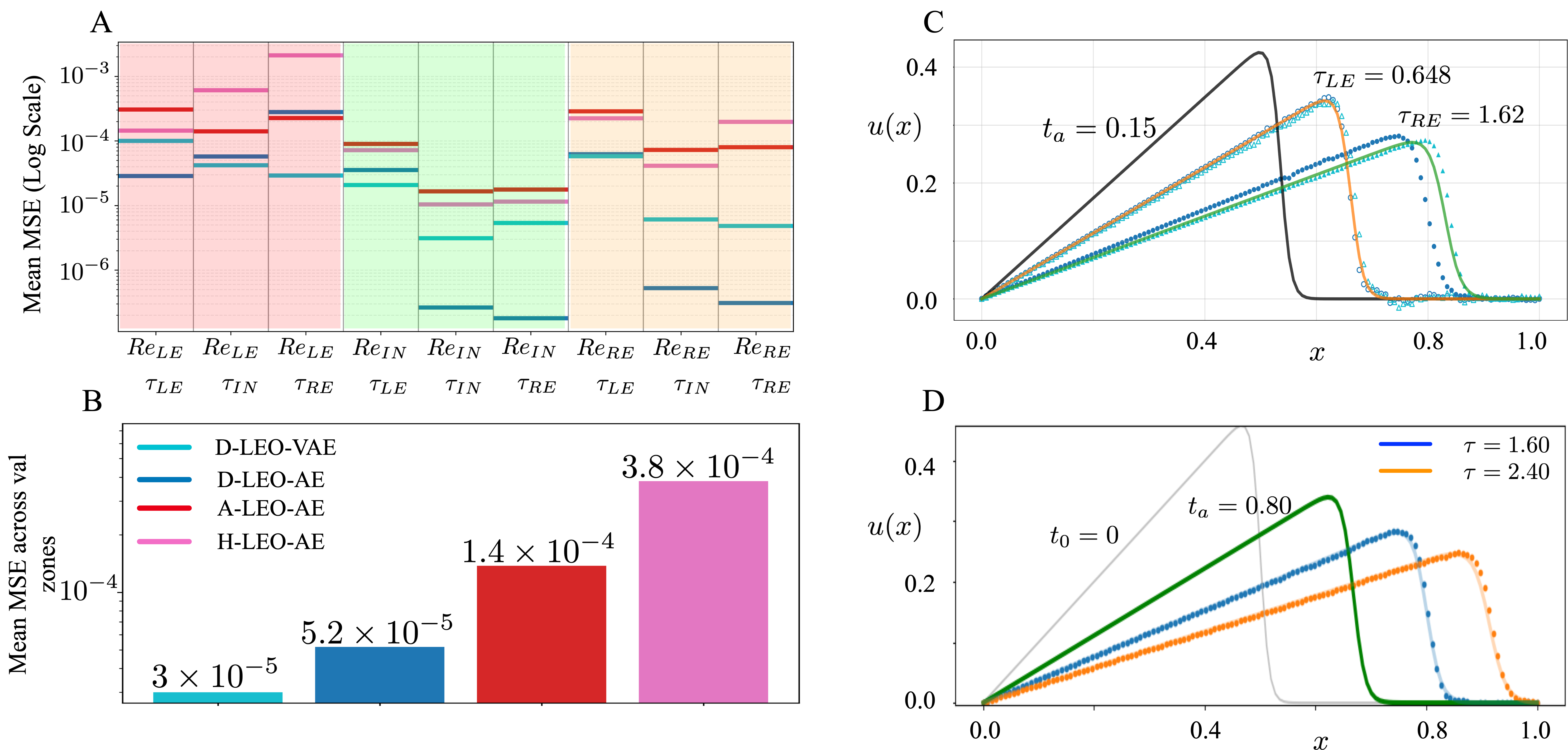}
  \caption{\textbf{Direct forecasting performance for Burgers' equation.}
  \textbf{(A)} Zone-wise mean-squared error (MSE, log scale) across the $3\times 3$ grid of $(\mathrm{Re},\tau)$ validation zones combining left/interpolation/right ranges in each variable.
  \textbf{(B)} Mean MSE averaged over all nine zones for each LEO variant.
  \textbf{(C)} High-dimensional forecasts for the two best models evaluated in their worst-performing low-$\mathrm{Re}$ extrapolation configuration: the black curve is the anchor state $u(t_a)$ with $t_a=0.15$, solid curves are exact solutions at two future horizons $\tau_{\mathrm{LE}}$ and $\tau_{\mathrm{RE}}$, and scatter markers are the corresponding model predictions.
  \textbf{(D)} Anchor agnostic prediction: the grey curve is the initial state $u(0)$, the green curve is an anchor state at $t_a=0.80$, and colored solid curves/markers are exact solutions and predictions at the same two horizons, illustrating that LEO can start from arbitrary anchor states along the trajectory.}
  \label{fig:Burgers'_mse}
  
\end{figure}

We evaluate the four LEO variants in a direct forecasting setting, where the model receives an anchor state $\bu(t_a)$, the Reynolds number $\mathrm{Re}$, and a forecast horizon $\tau$, and predicts the future state $\bu(t_a+\tau)$. The construction of state-to-state tuples and the $(\mathrm{Re},\tau)$ train--validation partitioning follow \S~\ref{data_training} and \S~\ref{sec:train_val_zones}, respectively. The validation set is stratified into a $3\times 3$ grid of zones combining left/interpolation/right ranges in both $\mathrm{Re}$ and $\tau$ (Fig.~\ref{fig:Burgers'_mse}A). For each zone, we draw $10^3$ random tuples $(\mathrm{Re},\tau,t_a)$ within the corresponding ranges and report the mean squared error (MSE) in the original 128-dimensional state space.

Figure~\ref{fig:Burgers'_mse}A reports the zone-wise MSE on a logarithmic scale. All models achieve their lowest error in the fully interpolative zone $(\mathrm{Re}_{\mathrm{IN}},\tau_{\mathrm{IN}})$ and degrade as we move away from the training region. For this Burgers' configuration, the largest errors for all models occur in the low-Reynolds left-extrapolation regime $\mathrm{Re}_{\mathrm{LE}}$, where the dynamics are diffusion-dominated and the solution is particularly smooth.

Figure~\ref{fig:Burgers'_mse}B aggregates performance by averaging the error across all nine validation zones. By this metric, the D--LEO--VAE attains the lowest overall error, followed by the D--LEO--AE. The consistent and hybrid variants exhibit higher aggregate error under the present training setup, which reflects differences in supervision objectives (latent consistency versus direct forecasting) and, for the consistency-based models, the staged warm-up schedule for $\lambda_{\mathrm{con}}$. Based on this result, we adopt the Direct LEO with a VAE backbone (D--LEO--VAE) as the default model for the remainder of this work.

To visualize worst-case behavior, Fig.~\ref{fig:Burgers'_mse}C focuses on the poorest validation configurations of the two best-performing models. From the zone-wise MSE, D--LEO--VAE attains its largest error in $(\mathrm{Re}_{\mathrm{LE}},\tau_{\mathrm{LE}})$ and D--LEO--AE in $(\mathrm{Re}_{\mathrm{LE}},\tau_{\mathrm{RE}})$. We fix an anchor time $t_a = 0.15$ and select a Reynolds number in $\mathrm{Re}_{\mathrm{LE}}$, and plot forecasts at $\tau_{\mathrm{LE}}$ and $\tau_{\mathrm{RE}}$ sampled from these zones (here $\tau_{\mathrm{LE}}\approx 0.648$ and $\tau_{\mathrm{RE}}\approx 1.624$). The black curve shows the anchor state $\bu(t_a)$, and the solid curves denote the exact solutions at times $t_a+\tau_{\mathrm{LE}}$ and $t_a+\tau_{\mathrm{RE}}$. Circles denote predictions of the direct AE model, while triangles denote predictions of the direct VAE model. Even in these worst-performing regimes, both models closely track the shock location and amplitude, so the relatively small MSE values correspond to only modest visible discrepancies.

Because LEO is trained in a state-to-state fashion, it enables prediction from arbitrary anchor states $\bu(t_a,\mathrm{Re})$ rather than only from the initial condition. Figure~\ref{fig:Burgers'_mse}D illustrates this for a sample in the $\mathrm{Re}_{\mathrm{RE}}$ regime: the initial condition $\bu(0)$ is shown in grey, while the anchor state at $t_a=0.80$ is shown in green. Given this anchor and a forecast horizon $\tau$, the model predicts $\bu(t_a+\tau)$ at two future times, including one beyond the training window ($t\approx 2.40$). The predictions accurately track the advected structures, indicating that the learned latent evolution operator captures the underlying transport dynamics rather than relying on a fixed initial-time surrogate. Additional qualitative examples are provided in Appendix~\ref{appendix:Burgers'_qualitative}.

\subsubsection{Latent consistency and oracle factorization.}
\label{sec:latent_consistency}

\begin{figure}[t]
  \centering
  \includegraphics[width=\textwidth]{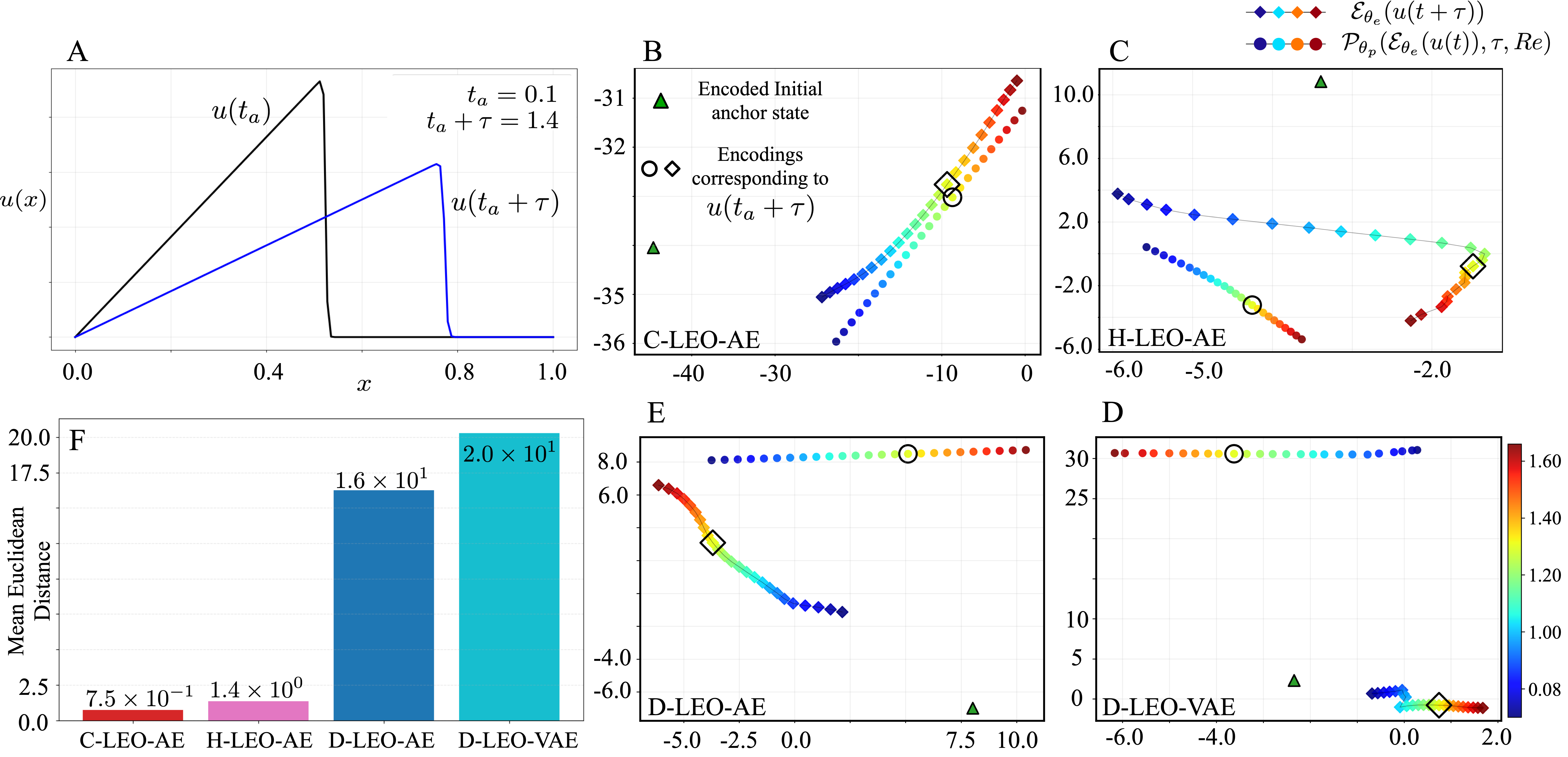}
  \caption{\textbf{Latent consistency across LEO variants.}
  Panel A: anchor state $\bu(t_a)$ and one representative future state $\bu(t_a+\tau)$, shown for a fixed Burgers' configuration with $t_a = 0.1$ and $\mathrm{Re}=2400$. For the latent trajectory analysis in Panels B--E, forecast horizons are varied over $\tau \in [0.7,1.7]$ in steps of $0.04$ and shows latent trajectories for C-LEO-AE, H-LEO-AE,
  D-LEO-AE, and D-LEO-VAE, respectively. In each panel, the green triangle marks the encoded anchor state, diamonds denote direct encodings of $\bu(t_a+\tau)$, and circles denote propagated latents $\hat \bz(t_a+\tau,\bzeta)$ for the same horizons; colors encode $\tau$. Panel F: mean Euclidean distance in latent space between encoded and propagated futures, averaged over randomly sampled $(\mathrm{Re},t_a,\tau)$ triples from the Burgers' validation set that decode to the same high-dimensional forecasted state.}
  \label{fig:leo_latent_consistency}
\end{figure}

To quantify how closely each LEO variant adheres to the oracle factorization in Theorem~\ref{thm:nn}, we compare the propagated latent
\[
\hat \bz(t+\tau,\mathrm{Re}) 
= \mathcal{P}_{\theta_p}\!\big(\tilde \bz(t,\mathrm{Re}),\tau,\mathrm{Re}\big)
\]
with the latent obtained by directly encoding the future state
\[
\tilde \bz(t+\tau,\mathrm{Re}) 
= \mathcal{E}_{\theta_e}\!\big(\bu(t+\tau,\mathrm{Re})\big),
\]
and define the latent consistency error
\[
d_{\mathrm{lat}}(t,\tau,\mathrm{Re}) 
= \big\|\hat \bz(t+\tau,\mathrm{Re}) - \tilde \bz(t+\tau,\mathrm{Re})\big\|_2.
\]
We evaluate $d_{\mathrm{lat}}$ across the Burgers' validation set, visualize representative latent trajectories, and aggregate statistics in Fig.~\ref{fig:leo_latent_consistency}.

Panel ~\ref{fig:leo_latent_consistency}A shows an anchor state $\bu(t_a)$ and one representative future state $\bu(t_a+\tau)$, shown for a fixed Burgers' configuration with $t_a = 0.1$ and $\mathrm{Re}=2400$. 
Panels~\ref{fig:leo_latent_consistency}B--E fix an anchor time $t_a = 0.1$ and Reynolds number $\mathrm{Re}=2400$, and evaluate horizons $\tau \in [0.7,1.7]$ with step size $0.04$. For each $\tau$, we compare the direct encoding
$\tilde \bz(t_a+\tau,\mathrm{Re}) = \mathcal{E}_{\theta_e}\!\big(\bu(t_a+\tau,\mathrm{Re})\big)$
with the propagated latent
$\hat \bz(t_a+\tau,\mathrm{Re}) = \mathcal{P}_{\theta_p}\!\big(\tilde \bz(t_a,\mathrm{Re}),\tau,\mathrm{Re}\big)$.

In each panel, the green triangle marks the encoded anchor state $\tilde \bz(t_a,\mathrm{Re})$, diamonds denote the directly encoded future states $\tilde \bz(t_a+\tau,\mathrm{Re})$, and circles denote the corresponding propagated latents $\hat \bz(t_a+\tau,\mathrm{Re})$; colors encode $\tau$. 

The consistent LEO (Panel~\ref{fig:leo_latent_consistency}B) produces nearly overlapping diamonds and circles along a single smooth curve, indicating that the encoder and propagator share a common latent chart consistent with the oracle relation. The hybrid model exhibits slightly larger but still moderate deviations. In contrast, the direct variants (AE and VAE backbones) place propagated latents on branches that are well separated from the direct encodings, yet both branches decode to nearly indistinguishable Burgers' profiles. This demonstrates that accurate forecasting does not require a unique latent representation: multiple latent configurations can be functionally equivalent under the decoder.

To summarize this behavior over the full validation set, we compute the mean latent consistency error $\mathbb{E}[d_{\mathrm{lat}}]$ over approximately $6000$ randomly sampled tuples $(\mathrm{Re},t_a,\tau)$ for each model. The results are shown in Fig.~\ref{fig:leo_latent_consistency}F. The consistent LEO achieves the smallest average distance, while the direct VAE variant exhibits the largest separation. 

Together with Fig.~\ref{fig:Burgers'_mse}, this reveals a fundamental trade-off between latent consistency and forecasting accuracy: the consistent and hybrid models most closely satisfy the oracle factorization, whereas the direct VAE achieves the lowest prediction error by allowing multiple functionally equivalent latent representations.

\subsection{Data scaling law and latent interpretation for D-LEO-VAE}
\label{sec:interpretability}

Disentangled latent representations are important for generalization in dynamical-systems modeling~\citep{Bengio2012,Fotiadis2023}. We assess how the D--LEO--VAE model scales with data and whether its latent space admits a simple physical interpretation.

\begin{figure}[t]
    \centering
    \includegraphics[width=\textwidth]{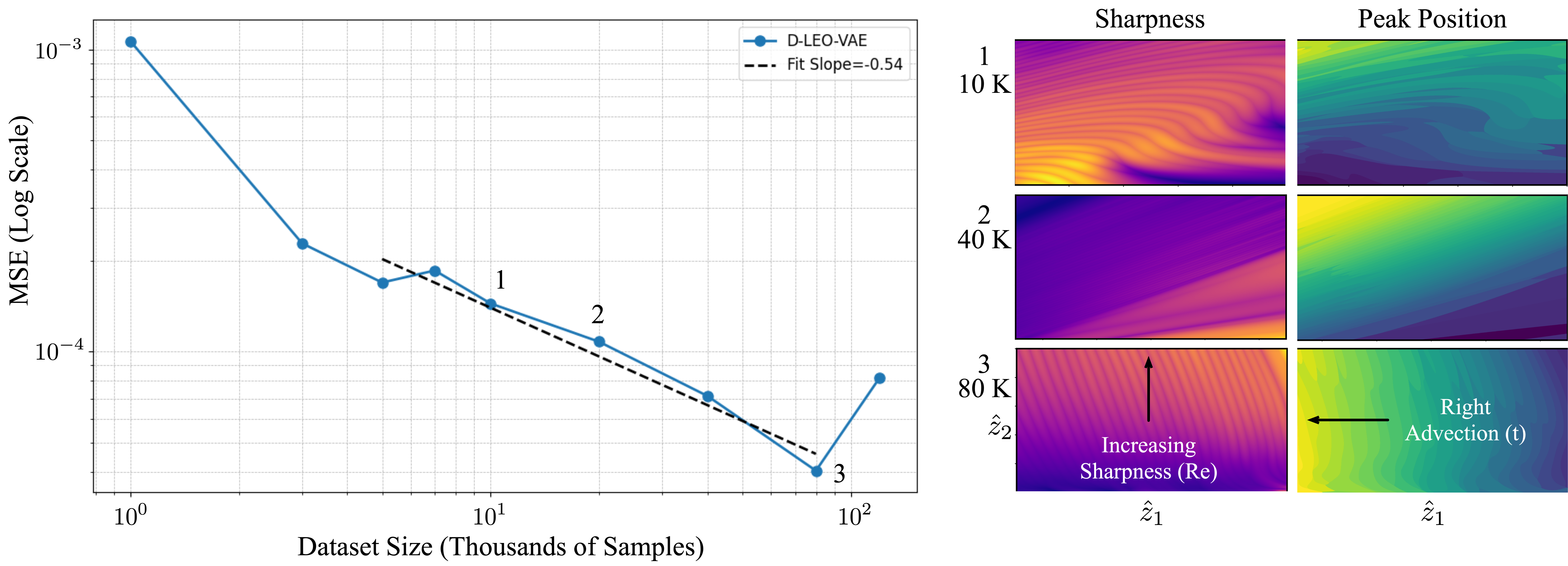}
    \caption{
    \textbf{Scaling and latent structure of D-LEO-VAE on Burgers'.}
    \emph{Left:} Mean squared error (MSE, log scale) in the four joint
    extrapolation zones obtained by combining the left/right extrapolation ranges of $\mathrm{Re}$ and $\tau$, averaged over randomly sampled triples $(\mathrm{Re},\tau,t_a)$ in each zone, as a function of dataset size (number of state--to--state training tuples, log scale). Between roughly $7\text{k}$ and $80\text{k}$ tuples the error follows an approximate power law with slope $\approx -0.54$, indicating data-efficient generalization in the most extrapolative regimes. \emph{Right:} Latent-space structure for three training sizes (rows 1--3). Each point $(\hat z_1,\hat z_2)$ in a uniform grid is decoded to a high-dimensional field, from which we extract sharpness ($\max_x |\partial u / \partial x|$; left column) and peak position (right column).
    }
    \label{fig:scaling_disentangle}
\end{figure}

We train D--LEO--VAE on Burgers' datasets with varying numbers of state-to-state tuples and evaluate in the most challenging joint extrapolation regimes, where both the Reynolds number $\mathrm{Re}$ and the forecast horizon $\tau$ lie outside the training range. Specifically, we focus on the four joint extrapolation zones obtained by combining the left/right extrapolation ranges of $\mathrm{Re}$ and $\tau$. The left panel of Fig.~\ref{fig:scaling_disentangle} shows the resulting mean validation MSE (log scale), averaged over these joint extrapolation zones, as a function of dataset size (log scale). Between roughly $7\text{k}$ and $80\text{k}$ tuples, the error decays almost linearly on the log--log plot; a least-squares fit over this range yields a slope of approximately $-0.54$, corresponding to the empirical scaling law $\mathrm{MSE} \sim p^{-0.54}$,
where $p$ denotes the number of training tuples. This monotone trend and approximate power-law regime indicate good data efficiency and stable generalization even in the most extrapolative $(\mathrm{Re},\tau)$ settings.

To analyze latent structure, we decode a uniform grid in the two-dimensional propagated latent coordinates
$\hat{\bz}=(\hat z_1,\hat z_2)$
and compute two summary statistics from each decoded field:
\emph{sharpness}, defined as $\max_x |\partial u/\partial x|$, and
\emph{peak position}, defined as the location of $\max_x u$.
Empirically, sharpness correlates with viscosity through $\mathrm{Re}$, while peak position reflects a transport-like coordinate in the decoded solutions. The right panel of Fig.~\ref{fig:scaling_disentangle} visualizes these quantities over the decoded latent grid for three training-set sizes (rows 1--3). For D--LEO--VAE, the latent space is smooth and well structured: variation along $\hat z_1$ corresponds predominantly to shifts in peak position, whereas variation along $\hat z_2$ primarily controls sharpness. This disentanglement becomes more pronounced as the dataset size increases, suggesting that the propagator, together with the forecasting objective, encourages physically meaningful latent coordinates\footnote{We also tested a positional-encoding latent propagator (PEP) variant for D--LEO--VAE on the same Burgers' benchmark. It yielded consistently higher zone-wise MSE and less structured latent maps than the concatenation-based propagator; full results are provided in Appendix~\ref{appendix:pep_results}.}.

Finally, we probe the local geometry of the decoder around encoded and propagated latents for D--LEO--VAE (Appendix~\ref{app:decoder_geometry}). For pairs of codes $\tilde{\mathbf{z}}(t+\tau,\mathrm{Re})$ and $\hat{\bz}(t+\tau,\mathrm{Re})$ that decode to nearly identical forecasted states, we analyze the decoder Jacobian
$
J(\bz) = \frac{\partial \mathcal{D}_{\theta_d}(\bz)}{\partial \bz}$. 
We find that $\tilde{\bz}$ lies in regions with stronger local expansion and anisotropy, characterized by larger singular values and larger $\det(J^\top J)$, whereas $\hat{\bz}$ resides in smoother, better-conditioned regions. This suggests that the D--LEO--VAE propagator can steer trajectories toward geometrically favorable latent charts while maintaining accurate long-horizon forecasts.

\subsection{Comparison with latent neural operator benchmark}
\label{sec:comparison}

We compare LEO against the closest latent neural-operator baseline, Latent DeepONet (L-DeepONet). DeepONet maps between function spaces using a dual-network structure, where a branch network encodes the input function and a trunk network encodes the query coordinates; predictions are obtained via an inner product of branch and trunk features. L-DeepONet applies this mechanism in a compressed representation by first encoding the input field with an autoencoder and then learning the operator in latent space, making it the closest baseline to our LEO formulation. In contrast, LEO jointly learns the encoder, latent evolution operator, and decoder in a single end-to-end framework.

\noindent\textbf{Fixed initial condition (\texorpdfstring{$t_a=0$}{ta=0}) protocol.}

\begin{figure}[h!]
  \centering
  \includegraphics[width=\linewidth]{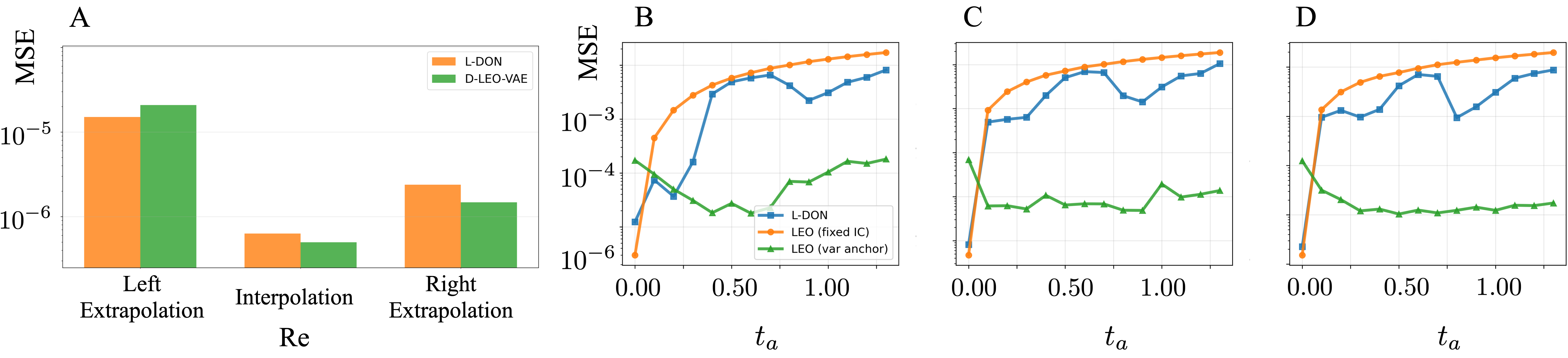}
  \caption{\textbf{Comparison with latent neural operators and anchor-time generalization.}
  \textbf{A}: Fixed-IC ($t_a=0$) performance across Reynolds-number regions for L-DeepONet and D-LEO-VAE; bars show mean MSE on a logarithmic $y$-axis.
  \textbf{B}--\textbf{D}: Temporal generalization across varying anchor times. Mean MSE versus anchor time $t_a$ for three Reynolds regimes, comparing L-DeepONet, D-LEO-VAE trained in the fixed-IC regime, and D-LEO-VAE trained on varying anchor states. For each $t_a$, we draw $500$ random pairs $(\mathrm{Re},\tau)$ from the indicated regime and report the mean test MSE, enforcing $t_a+\tau\le 2$ to remain within the simulation time window. L-DeepONet and fixed-IC D-LEO-VAE both experience anchor-time distribution shift for $t_a>0$, whereas the varying-anchor D-LEO-VAE remains accurate across the full range.
  \textbf{B}: left-extrapolation $\mathrm{Re}\in[400,600]$;
  \textbf{C}: interpolation $\mathrm{Re}\in[1006,1206]$;
  \textbf{D}: right-extrapolation $\mathrm{Re}\in[2200,2400]$.}
  \label{fig:op_comparison}
\end{figure}

To enable an architecture-level comparison with operator models trained from an initial condition, we retrain D-LEO-VAE in a fixed-IC setting. The anchor time is fixed to $t_a = 0$, and each training tuple has the form
\[
\bigl(u(x,0,\mathrm{Re}),\; u(x,t,\mathrm{Re}),\; t,\; \mathrm{Re}\bigr),
\]
where $t\in[0,2]$ is the absolute query time. Both models are therefore trained to map the initial state at $t=0$ to the solution at time $t$. Training uses only the in-distribution Reynolds bands $\mathrm{Re}_{\mathrm{TR_1}}$ and $\mathrm{Re}_{\mathrm{TR_2}}$, while evaluation follows the left-extrapolation, interpolation, and right-extrapolation regimes of $\mathrm{Re}$. For each regime, we draw approximately $n=1000$ random pairs $(\mathrm{Re},t)$ and report the mean and standard deviation of the MSE over the spatial domain. Both models achieve comparable performance. Panel~A of Fig.~\ref{fig:op_comparison} and Table~\ref{tab:exp1_numbers} summarize these results.

\begin{table}[h!]
\centering
\small
\begin{tabular}{lcc}
\toprule
\textbf{$\mathrm{Re}$} & \textbf{L-DeepONet} & \textbf{D-LEO-VAE} \\
\midrule
LE  &
$\mathbf{1.51{\times}10^{-5}}\ \pm\ 1.82{\times}10^{-5}$ &
$2.09{\times}10^{-5}\ \pm\ 9.60{\times}10^{-5}$ \\
IN &
$6.30{\times}10^{-7}\ \pm\ 8.14{\times}10^{-7}$ &
$\mathbf{4.97{\times}10^{-7}}\ \pm\ 1.06{\times}10^{-7}$ \\
RE &
$2.38{\times}10^{-6}\ \pm\ 2.80{\times}10^{-6}$ &
$\mathbf{1.48{\times}10^{-6}}\ \pm\ 9.74{\times}10^{-7}$ \\
\midrule
Par.& $540{,}606$ & $486{,}198$ \\
\bottomrule
\end{tabular}
\caption{Fixed-IC ($t_a=0$) generalization over approximately $n{=}1000$ random $(\mathrm{Re},t)$ pairs per region. Entries report mean $\pm$ standard deviation of the MSE (lower is better) over the spatial domain. The last row lists the total number of trainable parameters in each model.}
\label{tab:exp1_numbers}
\end{table}

To keep capacities comparable, L-DeepONet uses the same encoder--decoder backbone as D-LEO-VAE; the total number of trainable parameters is reported in the last row of Table~\ref{tab:exp1_numbers}.

\noindent\textbf{Temporal generalization across anchor times.}

To assess temporal flexibility, we evaluate three models across anchor times $t_a\in[0,1.3]$: (i) L-DeepONet, (ii) D-LEO-VAE trained in the fixed-IC regime above, and (iii) D-LEO-VAE trained using the varying-anchor state-to-state formulation introduced in this work. For each $t_a$, we draw $500$ random pairs $(\mathrm{Re},\tau)$ within each of the three test $\mathrm{Re}$ regimes and report the mean MSE over horizons satisfying $t_a+\tau\le 2$. Panels~B--D of Fig.~\ref{fig:op_comparison} plot the resulting error curves.

For the varying-anchor D-LEO-VAE, the model receives the anchor state $\bu(t_a)$ together with $(\tau,\mathrm{Re})$ and predicts $\bu(t_a+\tau)$ directly. For L-DeepONet and the fixed-IC D-LEO-VAE, using $t_a>0$ induces an anchor-time distribution shift because these models are trained in the canonical initial-condition regime (inputs concentrated at $t_a=0$). This distinction is clearly reflected in Panels~B--D: both L-DeepONet and fixed-IC D-LEO-VAE exhibit substantially larger errors as the anchor moves away from $0$, whereas the varying-anchor D-LEO-VAE maintains consistently low error across the full anchor-time range in all three Reynolds regimes.

This comparison highlights that the key advantage of LEO lies not only in its latent operator parameterization, but also in its training protocol: learning directly from state-to-state tuples across multiple anchor times enables strong anchor-agnostic generalization, whereas fixed-IC training leads to degradation under anchor-time shift.

For completeness, we compare query-time runtime against an auto-regressive baseline (AE--LSTM). AE--LSTM exhibits approximately linear growth in runtime with $\tau$ due to sequential $\mathcal{O}(\tau)$ rollouts, while D-LEO-VAE maintains essentially constant $\mathcal{O}(1)$ cost across offsets. D-LEO-VAE outperforms AE--LSTM by up to $50\times$ on CPU and $90\times$ on GPU (Appendix~\ref{app:ae_lstm_runtime}).

\section{Case studies on higher-dimensional PDEs}

To assess scalability and generalization in more complex spatiotemporal settings, we evaluate the proposed framework on higher-dimensional PDE systems, focusing on its ability to learn compact latent representations and enable accurate state-to-state forecasting.

\subsection{2D Advection--Diffusion}
\label{2d_advection_diffusion}

The 2D advection--diffusion equation models the transport of a scalar field
\(u(x,y,t)\) (e.g., temperature or concentration) under the combined effects of advection and isotropic diffusion:
\begin{equation} \label{eq:pde}
\frac{\partial u}{\partial t}
+ \frac{\partial u}{\partial x}
= \frac{1}{\mathrm{Re}} \left(
\frac{\partial^2 u}{\partial x^2}
+ \frac{\partial^2 u}{\partial y^2}
\right),
\end{equation}
where \(\mathrm{Re} = cL/\nu\), with $c$ denoting the constant advection velocity in the \(x\)-direction and \(\nu\) the diffusion coefficient. The left-hand side captures temporal evolution and directional transport, while the right-hand side represents diffusive spreading.

Assuming an initial point source at the origin, $u(x,y,0) = \delta(x,y)$, and neglecting boundary effects (i.e., an infinite domain), the solution admits the closed form
\begin{equation} \label{eq:solution}
u(x,y,t)
= \frac{1}{4 \pi \nu t}
\exp\!\left(
-\frac{(x - c t)^2 + y^2}{4 \nu t}
\right),
\qquad t > 0.
\end{equation}
This solution corresponds to a Gaussian pulse advected along the \(x\)-direction at speed \(c\), while diffusing isotropically in space. We use this analytical form solely to generate reference data; the model itself is trained purely on observed trajectories without access to the governing equation or its solution.

For the numerical experiments, we set \(c = 1\) and discretize the solution on a \(128 \times 128\) uniform grid over the domain \(-2L \le x \le 2L\), \(-2L \le y \le 2L\), with $L = 1$ and final time $T = 2$. The time step is fixed at \(\Delta t = 0.004\), and forecasts are evaluated over offsets \(\tau \in [0.6, 1.7]\).

To parameterize the relative strength of advection and diffusion, we vary \(\mathrm{Re} \in [1,10]\), where smaller values correspond to diffusion-dominated dynamics and larger values to advection-dominated transport. As in the Burgers'' experiments, we adopt an extrapolative validation protocol: training is performed on a central band of \((\mathrm{Re}, \tau)\), while evaluation probes left-extrapolation, interpolation, and right-extrapolation regions in the \((\mathrm{Re},\tau)\) plane.

\noindent\textbf{Model architecture and performance.}

Guided by the Burgers' experiments, we employ only the direct latent evolution operator with a variational autoencoder backbone (D-LEO-VAE) for this system. A latent dimension of $m=3$ is used, and the details on latent-dimension selection and hyperparameter tuning are provided in Appendix~\ref{app:2d_adv_diff_architecture}.

Figure~\ref{fig:axis1_adv_dif} summarizes model performance across the $(\mathrm{Re},\tau)$ space. We partition the domain into left-extrapolation, interpolation, and right-extrapolation bands in each variable, forming a $3\times3$ grid of validation zones. Axis~1 corresponds to low-\(\mathrm{Re}\) (diffusion-dominated), Axis~2 to interpolative regimes, and Axis~3 to high-\(\mathrm{Re}\) (advection-dominated) extrapolation.

Panels~A and~B show representative predictions along Axis~1 and Axis~3. For each axis, a single Re number is fixed within the corresponding band, while $\tau$ is varied across its three regimes. The anchor state is held fixed at $t_a = 0.1$, isolating the effect of forecast horizon.

In the diffusion-dominated regime (Axis~1), the model accurately captures smooth, isotropic spreading despite extrapolating beyond the training distribution. In the advection-dominated regime (Axis~3), it preserves sharp transported structures and their gradual diffusion across increasing $\tau$.

Panel~C reports the mean MSE (log scale) across all validation zones. For each $(\mathrm{Re},\tau)$ region, we sample $3000$ triplets $(t_a, \mathrm{Re}, \tau)$ with $t_a \in [0,2]$, and average the spatial error.

Consistent with the Burgers' case, the lowest errors occur in the interpolative regime. In contrast to Burgers', however, errors increase in the right-extrapolation region corresponding to strongly advection-dominated dynamics. A single corner, $(\mathrm{Re}_{\mathrm{RE}},\tau_{\mathrm{LE}})$, exhibits elevated error, corresponding to sharply advected structures at short forecast horizons. This behaviour is consistent across seeds and latent dimensions. Importantly, absolute errors remain below \(10^{-4}\), indicating accurate predictions throughout the parameter space.

The smooth performance across all regimes highlights the ability of the D-LEO-VAE to embed physical parameters within the latent dynamics, enabling robust generalization across both diffusive and advective transport regimes.

\noindent\textbf{Mass conservation.}

To assess whether the learned evolution operator approximately preserves global mass, we integrate the scalar field over the spatial domain at the forecast time $t_a+\tau$ and compare the total mass of the prediction with that of the reference solution. On the discrete grid, the mass is approximated by the quadrature
\[
M(t_a+\tau) \approx \sum_{i,j} u(x_i,y_j,t_a+\tau)\,\Delta x\,\Delta y.
\]
For each $(\mathrm{Re},\tau)$ validation zone, we draw $3000$ random triplets $(t_a,\mathrm{Re},\tau)$ with $t_a\in[0,2]$ and $\tau$ sampled from the corresponding interval, and compute the mass ratio $\frac{M_{\mathrm{pred}}}{M_{\mathrm{exact}}}$. Panel~D of Fig.~\ref{fig:axis1_adv_dif} reports the zone-wise mean mass ratio. Across all nine zones, this ratio remains close to unity, typically within a few percent and never deviating by more than approximately $6\%$. This indicates that the model preserves total scalar mass to good accuracy even though no explicit conservation constraint is imposed during training.

\begin{figure}[t]
    \centering
    \includegraphics[width=\textwidth]{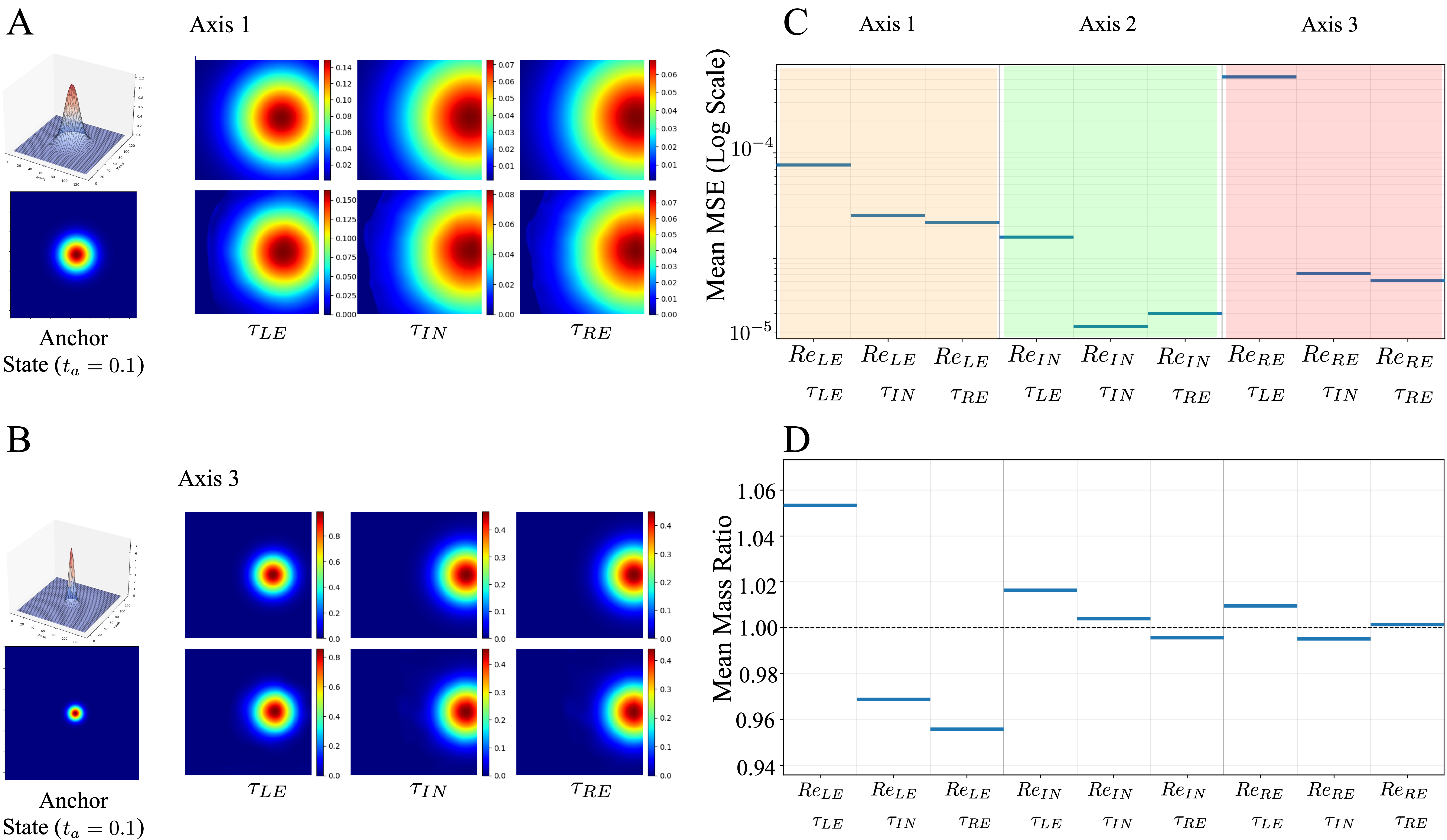}
    \caption{\textbf{2D advection--diffusion: latent evolution performance.}
    Here, Axis~1, Axis~2, and Axis~3 denote representative slices through the validation zones at fixed Reynolds-number bands ($Re_{\mathrm{LE}}$, $Re_{\mathrm{IN}}$, and $Re_{\mathrm{RE}}$), with $\tau$ varied across $\tau_{\mathrm{LE}}$, $\tau_{\mathrm{IN}}$, and $\tau_{\mathrm{RE}}$.
    \textbf{A}: Anchor state and representative forecasted states along Axis~1 (low-$Re$ slice, i.e., $Re_{\mathrm{LE}}$); top row: reference, bottom row: D-LEO-VAE prediction.
    \textbf{B}: Anchor state and representative forecasted states along Axis~3 (high-$Re$ slice, i.e., $Re_{\mathrm{RE}}$); top row: reference, bottom row: D-LEO-VAE prediction.
    \textbf{C}: Mean MSE (log scale) across all $(Re,\tau)$ validation zones, computed from $3000$ random $(t_a,Re,\tau)$ triplets per region, with $t_a+\tau\le 2$ and $\tau$ sampled from the corresponding $\tau$ interval.
    \textbf{D}: Mean mass ratio $M_{\mathrm{pred}}/M_{\mathrm{exact}}$ in each $(Re,\tau)$ validation zone, showing that the learned evolution operator preserves total scalar mass to within a few percent across all regions.}
    \label{fig:axis1_adv_dif}
\end{figure}

\vspace{0.7em}
\noindent\textbf{Comparison with Latent DeepONet: temporal generalization.}

\begin{figure}[t]
  \centering
  \includegraphics[width=\textwidth]{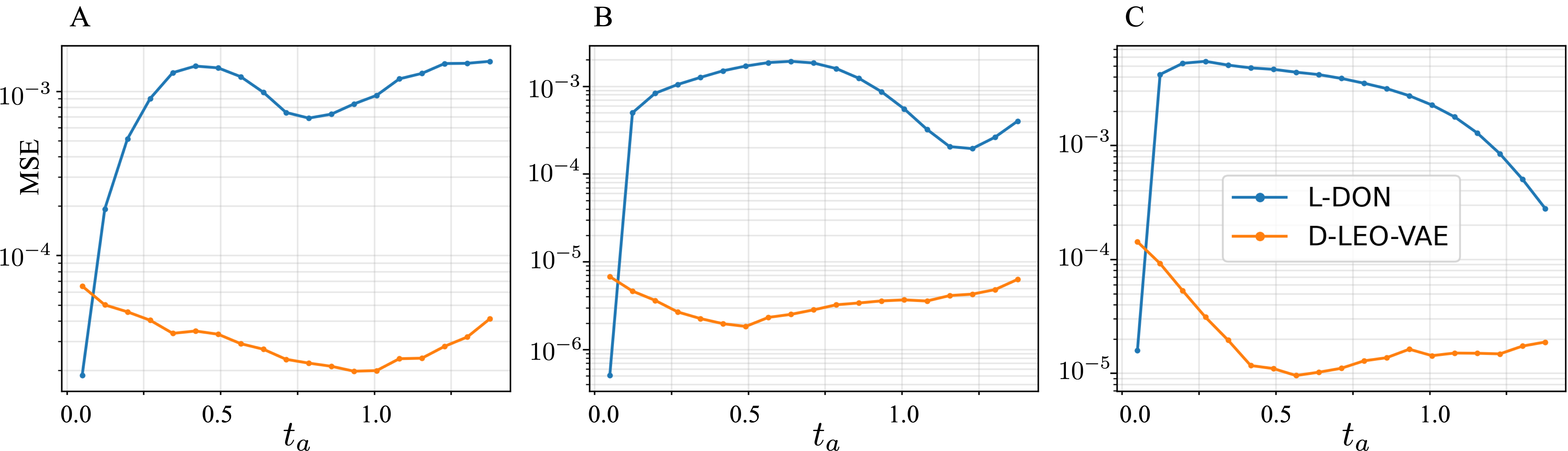}
  \caption{\textbf{2D advection--diffusion: temporal generalization.}
  Mean MSE versus anchor time $t_a$ for L-DeepONet and D-LEO-VAE across three Reynolds regimes:
  left extrapolation, interpolation, and right extrapolation.
  For each $t_a$, we sample $1000$ random pairs $(Re,\tau)$ per regime and average over feasible horizons with
  $t_a+\tau< 2 $.}
  \label{fig:advdiff_t0_sweep}
\end{figure}

We compare D-LEO-VAE against Latent DeepONet (L-DeepONet) under a state-to-state evaluation where the input is taken from arbitrary initial times $t_a$ rather than from a fixed initial condition. For each $t_a$ in a uniform sweep over $[0.05,\,1.45)$, and for each Reynolds regime (left-extrapolation, interpolation, and right-extrapolation), we draw $1000$ random pairs $(Re,\tau)$ from the corresponding $(Re,\tau)$ band and compute the mean MSE over all valid forecasts satisfying $t_a+\tau < 2$. Figure~\ref{fig:advdiff_t0_sweep} reports the resulting error curves: panel~A corresponds to the left-extrapolation Re regime, panel~B to the interpolative Re band, and panel~C to the right-extrapolation regime.

Across all regimes, D-LEO-VAE exhibits consistently low error with weak dependence on $t_a$, indicating robust temporal generalization from arbitrary anchor states. In contrast, L-DeepONet incurs significantly larger errors and shows strong sensitivity to $t_a$, consistent with distribution shift away from the fixed initial-condition setting. Overall, D-LEO-VAE provides substantially more reliable state-to-state forecasting across both diffusion- and advection-dominated regimes.

\begin{table}[t]
\centering
\caption{Zone-aggregated MSE (mean $\pm$ std) for L-DeepONet and D-LEO-VAE over temporal generalization trials. For each regime we aggregate all sampled triplets $(t_a, Re,\tau)$ used in the $t_a$-sweep evaluation.}
\label{tab:advdiff_ldon_vs_dleo_t0}
\begin{tabular}{lcc}
\hline
\textbf{Re} & \textbf{L-DeepONet} & \textbf{D-LEO-VAE} \\
\hline
LE  & $9.96\times 10^{-4} \pm 6.74\times 10^{-4}$ & $3.19\times 10^{-5} \pm 3.69\times 10^{-5}$ \\
IN       & $9.51\times 10^{-4} \pm 6.56\times 10^{-4}$ & $3.63\times 10^{-6} \pm 4.94\times 10^{-6}$ \\
RE & $3.04\times 10^{-3} \pm 2.03\times 10^{-3}$ & $2.70\times 10^{-5} \pm 1.30\times 10^{-4}$ \\
\hline
\textbf{Par.} & \textbf{1,011,236} & \textbf{962,314} \\
\hline
\end{tabular}
\end{table}

\subsection{Shallow water equation}

We consider the 2D shallow water system~\citep{gupta2023towards}, implemented using the \textsc{SpeedyWeather} library~\citep{klower2022milankl}. This setup models a stochastically forced, barotropic atmosphere in a simplified climate-like regime. The governing equations are

\begin{equation}
\begin{aligned}
\frac{\partial \omega}{\partial t}
+ \nabla \cdot \bigl( \boldsymbol{u}(\omega + f) \bigr)
&= F_{\omega} + \nabla \times \mathbf{F}_{\boldsymbol{u}}
 & \\  + (-1)^{n+1} \nu \nabla^{2n} \omega, \\
\frac{\partial \mathcal{D}}{\partial t}
- \nabla \times \bigl( \boldsymbol{u}(\omega + f) \bigr)
&= F_{\mathcal{D}} + \nabla \cdot \mathbf{F}_{\boldsymbol{u}}
   \\ & - \nabla^{2}\!\left(\tfrac{1}{2}(u^{2} + v^{2}) + g \eta\right)
   \\ & + (-1)^{n+1} \nu \nabla^{2n} \mathcal{D}, \\
\frac{\partial \eta}{\partial t}
+ \nabla \cdot (\boldsymbol{u} h)
&= F_{\eta},
\end{aligned}
\end{equation}
where $\boldsymbol{u}=(u,v)$ is the velocity field, $\omega$ is the vorticity, $g$ is gravitational acceleration, $f$ is the Coriolis parameter, $\eta$ is the interface height, $h$ is the layer thickness, and $\mathcal{D}=\nabla\cdot\boldsymbol{u}$ is the velocity divergence. The forcing terms $F_{\omega}$, $F_{\mathcal{D}}$, $F_{\eta}$, and $\mathbf{F}_{\boldsymbol{u}}=(F_u,F_v)$ introduce stochastic variability (see the \textsc{SpeedyWeather} documentation for details).

In our experiments, we focus only on the zonal velocity component $u$ as the forecasted field. To guide the choice of latent dimension, we estimate the intrinsic dimension of the data using the maximum-likelihood estimator \citep{campadelli2015intrinsic}, implemented in the \texttt{scikit-dimension} package. Based on this analysis, we use a latent dimension of $m=20$. Additional details on dataset construction and intrinsic-dimension estimation are provided in Appendix~\ref{app:additional_swe_details}.

Figure~\ref{fig:swe_single_shot_ic} illustrates single-shot forecasting when the anchor state coincides with the initial condition. We select a random test trajectory and use its snapshot at $t_a=0$ as input. From this single state, the model predicts the zonal velocity field at three lead times corresponding to increasing temporal offsets ($\tau \approx 3.4, 10.1, 19.5$ days). The top row shows the ground truth and the bottom rows shows the corresponding one-shot forecasts. The model qualitatively captures the dominant large-scale structures, and even for the longest horizon, the main jets and vortices remain well preserved, with discrepancies confined to fine-scale filaments. Additional examples (Appendix~\ref{app:swe_multi_anchor}) show forecasts from multiple anchor states along the same trajectory, confirming that the learned evolution operator generalizes beyond a fixed initial condition.

\begin{figure}[t]
  \centering
  \includegraphics[width=\textwidth]{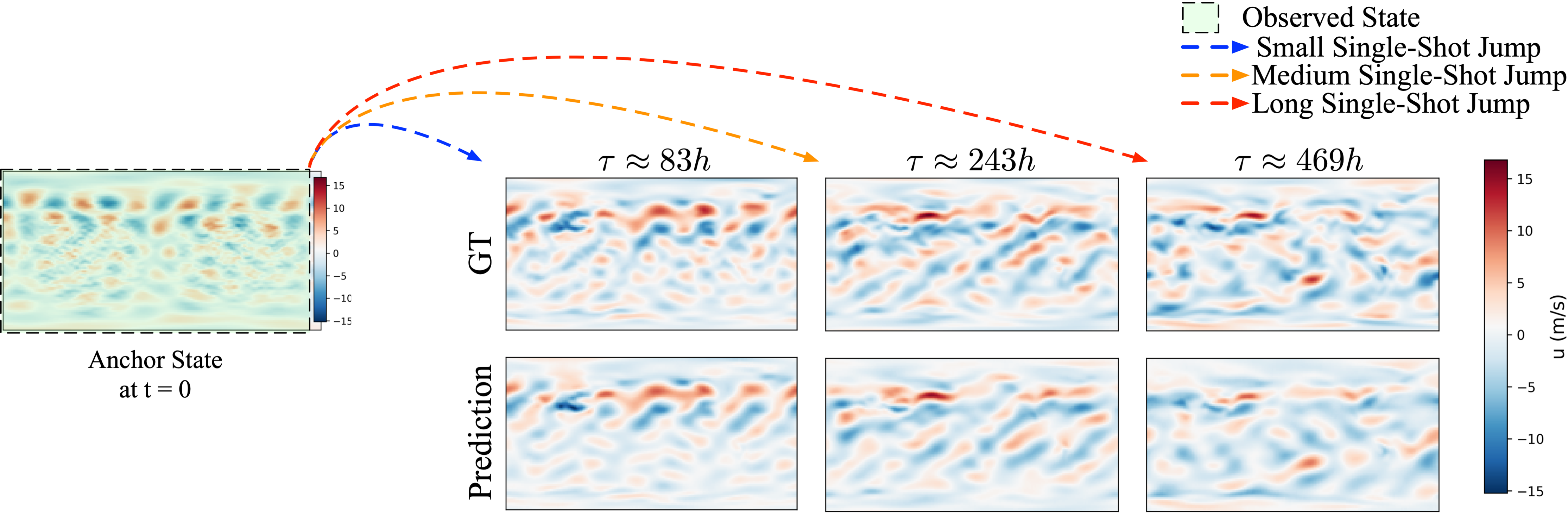}
  \caption{\textbf{Single--shot forecasts on the shallow--water benchmark.}
   One randomly selected test trajectory. The left panel shows the observed zonal velocity field at $t=0$. The three columns show ground truth (top) and LEO one--shot forecasts (bottom) for small, medium, and large lead times (approximately $3.4$, $10.1$, and $19.5$ days).}
  \label{fig:swe_single_shot_ic}
\end{figure}

\begin{figure}[t]
  \centering
  \includegraphics[width=\textwidth]{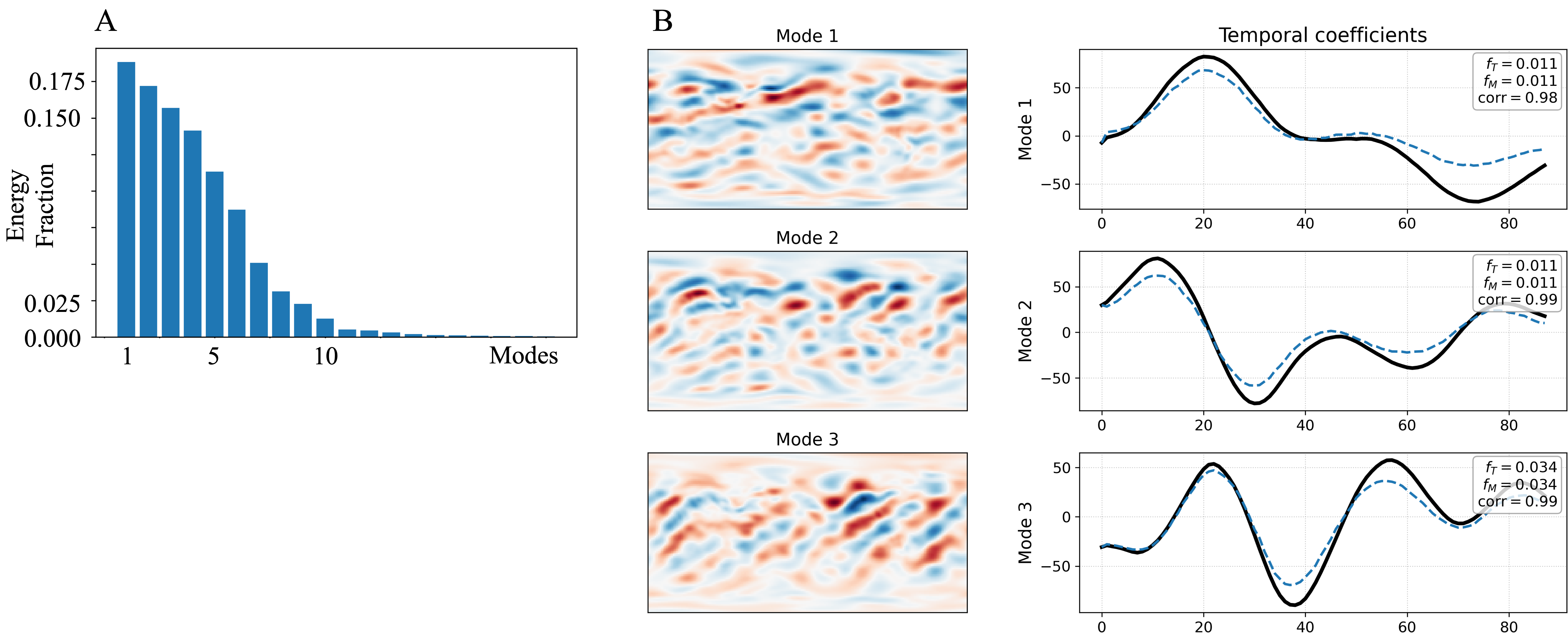}
  \caption{\textbf{POD-based comparison of SWE dynamics.}
  (A) Energy fraction of the leading POD modes and (B) Top three spatial POD modes (left) and their temporal coefficients for truth (solid black) and LEO forecasts projected onto the same basis (dashed blue). For each mode we report the dominant frequencies of the true and model coefficients ($f_T$, $f_M$) and their correlation,
  quantifying how well the learned evolution operator reproduces the large-scale dynamics.}
  \label{fig:swe_pod_dynamics}
\end{figure}

To further assess dynamical fidelity, we perform a proper orthogonal decomposition (POD) analysis on a representative test trajectory of the zonal velocity field. We compute POD modes from the ground-truth trajectory and project both the true fields and the model predictions onto this basis. Panel~\ref{fig:swe_pod_dynamics}(A) show that the leading $8$--$10$ modes capture more than $90\%$ of the variance, indicating strong low-dimensional structure. Panel~\ref{fig:swe_pod_dynamics}(B) compares the temporal coefficients of the first three modes. The model closely matches both amplitude and dominant frequency, where $f_T$ and $f_M$ denote the dominant frequencies of the true and model coefficients, respectively, estimated from the peak of the discrete Fourier spectrum after removing the mean, with correlations exceeding $0.98$. These results demonstrate that LEO not only reconstructs spatial structures, but also accurately captures their temporal evolution.

\section{Conclusion}

We presented the Latent Evolution Operator (LEO), a latent-space model for approximating finite-time evolution operators of parametric PDEs. The method maps a state at an arbitrary time, together with system parameters and a forecast horizon, to a future state in a single evaluation. Under standard manifold and regularity assumptions, finite-time evolution can be represented through an encoder--propagator--decoder structure. This enables learning evolution directly in latent space without recursive time integration. Numerical experiments on Burgers', advection--diffusion, and shallow-water systems show accurate predictions across parameters and time horizons, reduced computational cost relative to recurrent baselines, and improved performance when initialized from intermediate states compared to latent operator methods.

The current formulation assumes full-state observations and deterministic dynamics. Extensions to partially observed, noisy, and stochastic systems remain future work. Additional directions include incorporating physical constraints and applications to control and data assimilation. These results support learning finite-time evolution operators in latent space as a practical approach for modeling parametric dynamical systems.

\section*{Acknowledgments}

AGN acknowledges the support from the Department of Energy Early Career Research Award (Award no: DE-SC0022945) and the National Science Foundation AI Institute in Dynamic systems (Award no: 2112085, PM: Dr. Shahab Shojaei-Zadeh). WL acknowledges the support from NSF DMS 2145167 and 
DOE SC0024348.

\section*{CRediT authorship contribution statement}
\textbf{Khalid Rafiq:} Conceptualization, Methodology, Software, Investigation, Writing -- original draft.
\textbf{Wenjing Liao:} Methodology, Formal analysis, Writing -- review \& editing.
\textbf{Aditya G. Nair:} Conceptualization, Supervision, Methodology, Writing -- review \& editing, Funding acquisition.

\appendix 

\section{Theoretical guarantees for LEO}
 \label{app:theory}
 Our latent-evolution-operator (LEO) framework is based on a manifold hypothesis, which assumes that the set of PDE solutions $\{\boldsymbol{u}( t, {\bzeta})\} \subset \mathbb{R}^n$ lies on a compact, $m$-dimensional Riemannian manifold isometrically embedded in $\mathbb{R}^n$. After discretizing the PDE solution on a spatial grid with $n$ points, we denote $\boldsymbol{u}( t, {\bzeta}) \in \RR^n$ as
  the PDE solution vector parameterized by $\bzeta$ at time $t$.
 
 In this section, we provide a neural network representation theory about the supervised learning of the manifold structure and the latent propagator, which gives rise to a neural network representation guarantee of LEO.
 We will introduce preliminaries about manifold in Appendix \ref{appsub:pre}, present the operator decomposition theory in Appendix \ref{appsub:proofthmdecom}, and establish a neural network approximation theory in Appendix \ref{appsub:proofthmnn}. 

\subsection{Preliminary definitions and notation}
\label{appsub:pre}

 Let $\cM$ be a $m$-dimensional compact smooth Riemannian manifold isometrically embedded in $\RR^n$. We next introduce some definitions which can be found in \cite{tu2011manifolds} and \cite{lee2006riemannian}.

 \begin{definition}[Chart]
 A chart for $\cM$ is a pair $(U, \phi)$ such that $U \subset \cM$ is open and $\phi : U \mapsto \RR^m,$ where $\phi$ is a homeomorphism (i.e., bijective, $\phi$ and $\phi^{-1}$ are both continuous).
 \end{definition}
 The open set $U$ is called a coordinate neighborhood, and $\phi$ is called a coordinate map on $U$. A chart  defines a local coordinate system on $\cM$. We say two charts $(U, \phi)$ and $(V, \psi)$ on $\cM$ are $C^k$ compatible if and only if the transition functions,
 $\phi \circ \psi^{-1} : \psi(U \cap V) \mapsto \phi(U \cap V)$ and $\psi \circ \phi^{-1} : \phi(U \cap V) \mapsto \psi(U \cap V)$ are both $C^k$. We next give the definition of an atlas.
 \begin{definition}[$C^k$ Atlas]
 A $C^k$ atlas for $\cM$ is a collection $\{(U_\alpha, \phi_\alpha)\}_{\alpha \in \cA}$ of pairwise $C^k$ compatible charts such that $\bigcup_{\alpha \in \cA} U_\alpha = \cM$.
 \end{definition}

 \begin{definition}[Smooth Manifold] A smooth manifold is a manifold $\cM$ together with a $C^\infty$ atlas.
 \end{definition}
  The existence of an atlas on $\cM$ allows us to define differentiable functions.
 \begin{definition}[$C^k$ Functions on $\cM$]
 Let $\cM$ be a $m$-dimensional smooth manifold isometrically embedded in $\RR^n$. A function $f: \cM \mapsto \RR$ is $C^k$ for an atlas $\{U_\alpha,\phi_\alpha\}_
 {\alpha\in\cA}$, if for any $\alpha$, the composition $f \circ \phi_\alpha^{-1}: \phi(U_\alpha) \mapsto \RR$ is continuously differentiable up to order $k$.
 \end{definition}

 We next introduce the partition of unity, which plays a crucial role in our construction of encoder and decoder.
 \begin{definition}[Partition of Unity]
 A $C^\infty$ partition of unity on a manifold $\cM$ is a collection of non-negative $C^\infty$ functions $\rho_\alpha: \cM \mapsto [0,\infty)$ for $\alpha \in \cA$ such that
 \textbf{1)}. the collection of supports, $\{\textrm{supp} (\rho_\alpha)\}_{\alpha \in \cA}$ is locally finite\footnote{A collection $\{A_\alpha\}$ is locally finite if every point has a neighborhood that meets only finitely many of $A_\alpha$'s.}; \textbf{2)}. $\sum \rho_\alpha = 1$.
 \end{definition}
 For a smooth manifold, a $C^\infty$ partition of unity always exists.
 \begin{proposition}[Existence of a $C^\infty$ partition of unity]\label{thm:parunity}
 Let $\{U_\alpha\}_{\alpha \in \cA}$ be an open cover of a smooth manifold $\cM$. Then there is a $C^\infty$ partition of unity $\{\rho_i\}_{i=1}^\infty$ with every $\rho_i$ having a compact support such that $\textrm{supp}(\rho_i) \subset U_\alpha$ for some $\alpha \in \cA$.
 \label{prop:pou}
\end{proposition}

 To characterize the curvature of a manifold, we adopt the  reach of manifold introduced by Federer in \cite{federer1959curvature}.
 \begin{definition}[Reach, \emph{Definition 2.1} in \cite{aamari2019estimating}]
 Denote $\cC(\cM) = \{\bx \in \RR^n : \exists \boldsymbol{p} \neq \boldsymbol{q} \in \cM, \|{\boldsymbol{p} - \bx}\|_2 = \|{\boldsymbol{q} - \bx}\|_2 = \inf_{\boldsymbol y \in \cM} \|{\boldsymbol y- \bx}\|_2\}$ as the set of points that have at least two nearest neighbors on $\cM$. Then the reach $\tau_{\cM} $ is defined as $\tau_{\cM} := \inf_{\bx\in \cM, \boldsymbol y \in \cC(\cM)} \|{\bx - \boldsymbol y}\|_2.$
 \end{definition}

 Reach has a straightforward geometrical interpretation: At each point $\bx \in \cM$, the radius of the osculating circle is no less than $\tau_{\cM}$. 
 Reach determines a proper choice of an atlas for $\cM$. In Section \ref{appsub:proofthmnn}, we choose each chart $U_\alpha$ contained in a ball of radius less than $\tau_{\cM} / 4$. 

 \begin{definition}[Lipschitz function]
 An operator $f: \mathcal{X} \rightarrow \mathcal{Y}$ is $L$-Lipschitz if for all $u,v\in \mathcal{X}$, $\|f(u)-f(v)\|_{\mathcal{Y}}\le L \|u-v\|_{\mathcal{X}}$, where $\|\cdot\|_{\mathcal{X}}$ and $\|\cdot\|_{\mathcal{Y}}$ stand for the metric in $\mathcal{X}$ and $\mathcal{Y}$ respectively.
\end{definition}

 \noindent
 {\bf Notation.} We introduce some notation to be used in the proof of Theorem \ref{thm:nn}.
 For a function $f:\Omega \rightarrow \RR$, we denote its $L^\infty$ norm by $\|f\|_{L^\infty(\Omega)}:= \sup_{\bx\in\Omega}|f(\bx)|$.
 For two functions $f$ and $g$, we use $f\circ g$ to denote their composition. 

\subsection{Proof of Theorem \ref{thm:decom}}
\label{appsub:proofthmdecom}

Theorem \ref{thm:decom} is proved in a geometric encoder-decoder representation framework for the manifold $\cM$, followed by the establishment of a latent propagator.

  \noindent $\bullet$ {\bf A geometric encoder-decoder representation framework for the manifold $\cM$.} 
 We first present a geometric encoder-decoder representation framework for the manifold $\cM$, following \cite[Theorem 1]{chen2019efficient} and \cite[Lemma 4]{liu2024deep} and their proofs.
 We claim that, for the manifold $\cM$ in Assumption \ref{assumption:manifold}, there exist an oracle encoder and decoder pair $\cE^*:\cM\rightarrow \RR^{C_{\cM}(m+1)}$ and $\cD^*:{\rm Range}(\cE^*(\cM)) \rightarrow \cM$ satisfying 
 	\begin{align}
 		\cD^*\circ\cE^*(\bu)=\bu, \ \forall \bu\in \cM.
   \label{eq.ec}
	\end{align} 
     Here $C_{\cM}$ denotes the number of charts for $\cM$.
 	 We call $\cE^*$ and $\cD^*$ as the oracle encoder and decoder for the manifold $\cM$. These oracles are constructed in the proof of \cite[Theorem 1]{chen2019efficient} and \cite[Lemma 4]{liu2024deep}, but we still present its construction here for completeness.	The construction of $\cE^*$ and $\cD^*$ relies on a proper construction of an atlas of $\cM$ and a partition of unity of $\cM$. We construct $\cE^*$ and $\cD^*$ using the following two steps.
 
  \textbf{Step 1. Construction of an atlas}. Denote the open Euclidean ball with center $\bc$ and radius $r$ in $\RR^n$ by $\cB(\bc, r)$. For any $r>0$, the collection $\{\cB(\bc, r)\}_{\bc \in \cM}$ is an open cover of $\cM$. Since $\cM$ is compact, there exists a finite collection of points $\bc_i$ for $i = 1, \dots, C_\cM$ such that $\cM \subset \bigcup_i \cB(\bc_i, r)$.  

  We pick the radius $r < \tau_{\cM} / 4$ so that $U_i = \cM \cap \cB(\bc_i, r)$ is diffeomorphic\footnote{$P$ is diffeomorphic to $Q$ if there is a mapping $\Gamma : P \mapsto Q$ bijective, differentiable, and its inverse also being differentiable.} to a ball in $\RR^m$ \citep{niyogi2008finding}. For each $U_i$, we define $\phi_i$ to be the orthogonal projection to the tangent space of $\cM$ centered at $\bc_i$. According to \cite[Lemma 2]{chen2019efficient}, $\phi_i$ is a diffeomorphism \footnote{$\phi_i$ is differentiable and its inverse $\phi_i^{-1}$ is also differentiable}.
 We then let $\{(U_i, \phi_i)\}_{i=1}^{C_\cM}$ be an atlas on $\cM$. 
 The number of charts $C_\cM$ is upper bounded by
 $C_\cM \leq \left\lceil\frac{SA(\cM)}{r^m} T_m\right\rceil$
 where $SA(\cM)$ is the surface area of $\cM$, and $T_m$ is the thickness\footnote{Thickness is the average number of $U_i$'s that contain a point on $\cM$ \citep{conway2013sphere}.} of the $U_i$'s.
 The thickness $T_m$ scales approximately linear in the intrinsic dimension $m$ \citep{conway2013sphere}. 

 \textbf{Step 2. Construction of the oracle encoder and decoder}. For the atlas $\{(U_i, \phi_i)\}_{i=1}^{C_\cM}$, let $\{\rho_i\}_{i=1}^{C_\cM}$ be a $C^\infty$ partition of unity associated with $\{U_i\}_{i=1}^{C_\cM}$ such that for each $i$, $\rho_i$ is supported within $U_i$  and each $\rho_i$ is a $C^\infty$ function. The existence of such partition of unity is guaranteed by Proposition \ref{prop:pou} \citep{tu2011manifolds}.

 	We define the oracle encoder $\cE^*: \cM\rightarrow \RR^{C_{\cM}(d+1)}$ as
	\begin{equation}
 	\begin{aligned}
 		\cE^*(\bu)=\begin{bmatrix}
 			\bbf_1(\bu) & \cdots & \bbf_{C_{\cM}}(\bu)
 		\end{bmatrix}^{\top}, \\
 		\mbox{ with } \quad \bbf_j(\bu)=\begin{bmatrix}
 			(\phi_j(\bu))^{\top} & \rho_j(\bu)
 		\end{bmatrix}\in \RR^{m+1},
 		\label{eq:oracleencoder}
 	\end{aligned}
	\end{equation}
 	and the corresponding oracle decoder $\cD^*: \RR^{C_{\cM}(m+1)}\rightarrow \cM$ as
 	\begin{align}
 		\cD^*(\bz)=\sum_{j=1}^{C_{\cM}} \phi_j^{-1}((\bz_j)_{1:d})\times (\bz_j)_{d+1},
        \label{eq:oracledecoder}
 	\end{align}
 	where $(\bz_j)_{1:d}=\begin{bmatrix}
 		(\bz_j)_1 & \cdots & (\bz_j)_d 
 	\end{bmatrix}.$
 	For any $\bu\in \cM$, one can verify that
	\begin{equation*}
	\begin{aligned}
  \cD^*\circ\cE^*(\bu)=&\sum_{j=1}^{C_{\cM}} \phi_j^{-1}\circ\phi_j(\bu)\times \rho_j(\bu) \\
 		& =\sum_{\bu\in {\rm supp}(\rho_j)} \bu\times \rho_j(\bu) = \bu.
 	\end{aligned}
         \end{equation*}

  \noindent $\bullet$ {\bf Latent propagator.}   The PDE evolution from $\bu(t,\bzeta)$ to $\bu(t+\tau,\bzeta)$ is governed by the Lipschitz operator $\frakF$.
         Let $\bz(t,\bzeta) = \cE^*(\bu(t,\bzeta))$ and $\bz(t+\tau,\bzeta) = \cE^*(\bu(t+\tau,\bzeta))$ be the latent states. 

        With the oracle encoder $\cE^*$ and the oracle decoder $\cD^*$ on $\cM$, there exists an oracle latent propagator $\cP^*$ such that \eqref{eq:oracleprop} holds:
         \begin{equation*}
\bz(t+\tau,\bzeta) = 
 \cP^*(\bz(t,\bzeta),\tau,\bzeta) = \cE^*\circ\frakF(\cD^*(\bz(t,\bzeta)),\bzeta).
         \end{equation*}
         With the oracle propagator in \eqref{eq:oracleprop}, the PDE evolutionary operator $\frakF$ can be expressed as \eqref{eq:oracleevolve}:
         $$\bu(t+\tau,\bzeta) = \frakF(\bu(t,\bzeta),\bzeta)=\cD^*\circ\cP^*(\cE^*(\bu(t,\bzeta),\tau,\bzeta).$$
 A workflow about the PDE evolutionary operator $\frakF$, encoder $\cE^*$, decoder $\cD^*$ and latent propagator $\cP^*$ is illustrated in Figure \ref{fig:workflow}.

         Since the $\phi_j$'s and $\phi_j^{-1}$'s in \eqref{eq:oracleencoder} and \eqref{eq:oracledecoder} are differentiable, the oracle encoder $\cE^*$ and decoder $\cD^*$ are Lipschitz. We denote their Lipschitz constants by $\Lip_{\cE^*}$ and $\Lip_{\cD^*}$ such that $\|\cE^*(\bu_1)-\cE^*(\bu_2)\|\le \Lip_{\cE^*}\|\bu_1-\bu_2\|$ for $\bu_1,\bu_2\in\cM$ and $\|\cD^*(\bz_1)-\cD^*(\bz_2)\|\le \Lip_{\cD^*}\|\bz_1-\bz_2\|$ for $\bz_1,\bz_2 \in {\rm Range}(\cE^*(\cM))$. The Lipschitz constants  $\Lip_{\cE^*}$ and $\Lip_{\cD^*}$ depend on the derivatives of the coordinate maps $\phi_i$'s and the partition of unity $\rho_i$'s up to order $1$.
         We claim that the oracle propagator in \eqref{eq:oracleprop} is also Lipschitz: For $\bz_1(t,\bzeta),\tau,\bzeta)=\cE^*(\bu_1(t,\bzeta),\tau,\bzeta))$ and $\bz_2(t,\bzeta),\tau,\bzeta)=\cE^*(\bu_2(t,\bzeta),\tau,\bzeta))$, we have \begin{align*}
         &\|\cP^*(\bz_1(t,\bzeta),\tau,\bzeta)-\cP^*(\bz_2(t,\bzeta),\tau,\bzeta)\|
         \\
         = &\left\|\cE^*\circ\frakF(\bu_1(t,\bzeta),\bzeta) - \cE^*\circ\frakF(\bu_2(t,\bzeta),\bzeta)\right\|
         \\
          \le& \Lip_{\cE^*}\Lip_{\frakF}\|\bu_1-\bu_2\| \le \Lip_{\cE^*}\Lip_{\frakF}\Lip_{\cD^*}\|\bz_1-\bz_2\|.
         \end{align*}
        The Lipschitz constant of $\cP^*$ satisfies $\Lip_{\cP^*} \le \Lip_{\cE^*}\Lip_{\frakF}\Lip_{\cD^*}$, which depends derivatives of the coordinate maps $\phi_i$'s and $\phi_i^{-1}$'s and the partition of unity $\rho_i$'s up to order $1$.

 \subsection{Proof of Theorem \ref{thm:nn}}
 \label{appsub:proofthmnn}

 With the oracle encoder $\cE^*$, decoder $\cD^*$ and propagator $\cP^*$ given in Theorem \ref{thm:decom}, we will establish neural network approximation results for each of them, which gives rise to Theorem \ref{thm:nn}.
    
       \noindent $\bullet$ {\bf Encoder and decoder approximation theory.} 

  We next show that there exists networks $\cE_{\theta_e}$ and $\cD_{\theta_d}$ approximating $\cE^*$ and $\cD^*$ so that $\cD_{\theta_d}\circ\cE_{\theta_e}$ approximates $\cD^*\circ\cE^*$ with high accuracy. We will use the neural network approximation results in \cite{yarotsky2017error,chen2019efficient}.

 Lemma \ref{lemma:multiplication0} constructs a neural network to approximate the multiplication operation for bounded input.
 \begin{lemma}[Proposition 3 in \cite{yarotsky2017error}]\label{lemma:multiplication0}
 		For any $B>0$ and $0<\varepsilon<1$, there exists a network $\widetilde{\times}$ so that for any $|x_1|\leq B$ and $|x_2|\leq B$, we have
 		\begin{align*}
 			|\widetilde{\times}(x_1,x_2)-x_1\times x_2|<\varepsilon,\ \widetilde{\times}(x_1,0)=\widetilde{\times}(0,x_2)=0.
 		\end{align*}
 		Such a network has $O(\log \varepsilon^{-1})$ layers and parameters. The width is bounded by 6 and all parameters are bounded by $B^2$.
 	\end{lemma}

     Lemma \ref{lemma:Sobolev} constructs a neural network to universally approximate $C^k$ functions (these functions have derivatives up to order $k$ and all derivatives are bounded) on $[0,1]^d$.
      \begin{lemma}[Theorem 1 of \cite{yarotsky2017error}]
     \label{lemma:Sobolev}
     Let $M,F>0$.
         For any $\varepsilon\in(0,1)$, there is a network architecture $\cF_{\rm NN}(d,1,L,p,K,\kappa,M)$ with 
         \begin{equation*}
         \begin{aligned}
             &L=O(\log \varepsilon^{-1}), \ p=O(\varepsilon^{-d/k}), \\
             & \ K=O(\varepsilon^{-d/k}\log \varepsilon^{-1}), \ \kappa= O(\varepsilon^{-d/k})
         \end{aligned}
         \end{equation*}
         so that for any $C^k$ function $f^*:[0,1]^d\rightarrow[-M,M]$  with
     $\max_{\mathbf{k}: |\mathbf{k}| \le k} \sup_{\bx \in [0,1]^d} |D^{\mathbf{k}} f^*| \le F$, there exists a network $f_{\rm NN}\in \cF_{\rm NN}(d,1,L,p,K,\kappa,M)$ satisfying
         \begin{align*}
             \|f_{\rm NN}-f^*\|_{L^\infty([0,1]^d)}\leq \varepsilon.
         \end{align*}
         The constants hidden in $O(\cdot)$ depend on $d,k,M,F$.
     \end{lemma}

 Lemma \ref{lemma:chen} constructs a neural network to universally approximate $C^k$ functions (for an atlas, these functions have derivatives up to order $k$ with respect to the local coordinates, and all derivatives are upper bounded) on manifold.

     \begin{lemma}
     \label{lemma:chen}
 Let $\cM$ be the manifold in Theorem \ref{thm:nn}. Let $M,F>0$.
         For any $\varepsilon\in(0,1)$, there is a network architecture $\cF_{\rm NN}(n,1,L,p,K,\kappa,M)$ with 
         \begin{equation*}
         \begin{aligned}
             &L=O(\log \varepsilon^{-1}), \ p=O(\varepsilon^{-m/k}), \\ 
             & \ K=O(\varepsilon^{-m/k}\log \varepsilon^{-1}), \ \kappa= O(\varepsilon^{-m/k})
         \end{aligned}
         \end{equation*}
         so that for any $C^k$ function $f^*:\cM\rightarrow[-M,M]$  with
     $\max_{\mathbf{k}: |\mathbf{k}| \le k} \sup_{\bx \in [0,1]^d} |D^{\mathbf{k}} f^*| \le F$\footnote{These derivatives are defined on the coordinates for the atlas $\{U_i,\phi_i\}_{i=1}^{C_{\cM}}$.}, there exists a network $f_{\rm NN}\in \cF_{\rm NN}(n,1,L,p,K,\kappa,M)$ satisfying
        \begin{align*}
             \|f_{\rm NN}-f^*\|_{L^\infty(\cM)}\leq \varepsilon.
         \end{align*}
         The constants hidden in $O(\cdot)$ depend on $m,k,\tau_{\cM},F,M$, $B:=\sup_{\bx\in \cM}\|\bx\|_{\infty}$, $n$ (linear), and the derivatives of the coordinate maps $\phi_i$'s and the partition of unity $\rho_i$'s up to order $k$.
     \end{lemma}

 The encoder-decoder representation theory is guaranteed based on these neural network approximation lemmas.

\vspace{0.7em}
 {\bf Encoder approximation.}

For the oracle encoder $\cE^*$ in \eqref{eq:oracleencoder}, each $\phi_j$ is a linear projection, so it can be realized by a linear layer in neural network. Each $\rho_j$ is a $C^\infty$ function on $\cM$. We apply Lemma \ref{lemma:chen} considering $\rho_j \in C^{1}$ and $\rho_j \in [0,1]$. Then For any $\varepsilon_{\rho_j}\in(0,1)$, there is a network architecture $\cF_{\rm NN}(n,1,L_{\rho_j},p_{\rho_j},K_{\rho_j},\kappa_{\rho_j},1)$ with 
 \begin{equation*}
 \begin{aligned}
     &L_{\rho_j}=O(\log \varepsilon_{\rho_j}^{-1}), \ p_{\rho_j}=O(\varepsilon_{\rho_j}^{-m}), \\
     &\ K_{\rho_j}=O(\varepsilon_{\rho_j}^{-m}\log \varepsilon_{\rho_j}^{-1}), \ \kappa_{\rho_j}= O(\varepsilon_{\rho_j}^{-m})
 \end{aligned}
 \end{equation*}
 so that for the $\rho_j$ in \eqref{eq:oracleencoder}, there exists a network $\rho_{j,\rm NN}\in \cF_{\rm NN}(n,1,L_{\rho_j},p_{\rho_j},K_{\rho_j},\kappa_{\rho_j},1)$ satisfying
\begin{align*}
    \|\rho_{j,\rm NN}-\rho_j\|_{L^\infty(\cM)}\leq \varepsilon_{\rho_j}.
 \end{align*}
 The constants hidden in $O(\cdot)$ depend on $m,k,\tau_{\cM}$, $B:=\sup_{\bx\in \cM}\|\bx\|_{\infty}$, $n$ (linear), and the derivatives of the coordinate maps $\phi_i$'s and the partition of unity $\rho_i$'s up to order $1$.

 With each $\rho_j$ approximated by $\rho_{j,\rm NN}$ above, we can put the networks indexed by $j=1,\ldots,C_{\cM}$  in parallel, and then obtain a network approximating $\cE^*$ in \eqref{eq:oracleencoder}. Set $\varepsilon_{\rho_j}=\varepsilon_{E}$ for all $j$. Then there is a network architecture $\cF_{\rm NN}(n,{C_{\cM}}(m+1),L_{E},p_{E},K_{E},\kappa_{E},M_E)$ with 
 \begin{equation*}
\begin{aligned}
     &L_{E}=O(\log \varepsilon_{E}^{-1}), \ p_{E}=O(C_{\cM}\varepsilon_{E}^{-m}), \\ 
      & K_{E}=O(C_{\cM}\varepsilon_{E}^{-m}\log \varepsilon_{E}^{-1}), \ \kappa_{E}= O(\varepsilon_{E}^{-m}), \ M_{E}=O(1) 
 \end{aligned}
 \end{equation*}
 so that for the $\cE^*$ in \eqref{eq:oracleencoder}, there exists a network $\cE_{\theta_e}\in \cF_{\rm NN}(n,{C_{\cM}}(m+1),L_{E},p_{E},K_{E},\kappa_{E},M_E)$ satisfying
 \begin{align}
     \sup_{\bu \in \cM}\|\cE_{\theta_e}(\bu)-\cE^*(\bu)\|_{\infty}\leq \varepsilon_{E}.
     \label{eq:approxe}
 \end{align}
 The constants hidden in $O(\cdot)$ depend on $m,\tau_{\cM}$, $B:=\sup_{\bu\in \cM}\|\bu\|_{\infty}$, $n$ (linear), and the output, derivatives of the coordinate maps $\phi_i$'s and the partition of unity $\rho_i$'s up to order $1$.

\vspace{0.6em}
{\bf Decoder approximation.} For the oracle decoder $\cD^*$ in \eqref{eq:oracledecoder}, we first approximate each $\phi_j^{-1}: \phi_j(U_j) \rightarrow \cM\subset \RR^n$, which is differentiable. We apply Lemma \ref{lemma:Sobolev} for each output coordinate $(\phi_j^{-1})_k \in C^{1}$ for $k=1,\ldots,n$. For any $\varepsilon_{j,k}\in(0,1)$, there is a network architecture $\cF_{\rm NN}(m,1,L_{j,k},p_{j,k},K_{j,k},\kappa_{j,k},B)$ with 
\begin{equation*}
\begin{aligned}
    &L_{j,k}=O(\log \varepsilon_{j,k}^{-1}), \quad
    p_{j,k}=O(\varepsilon_{j,k}^{-m}),  \\
    &K_{j,k}=O(\varepsilon_{j,k}^{-m}\log \varepsilon_{j,k}^{-1}), \quad
    \kappa_{j,k}= O(\varepsilon_{j,k}^{-m}).
\end{aligned}
\end{equation*}
so that for the $(\phi_j^{-1})_k: \phi_j(U_j) \rightarrow [-B,B]$, there exists a network
\begin{equation*}
(\phi_j^{-1})_{k,\rm NN}
\in
\cF_{\rm NN}
\bigl(
m,1,L_{j,k},p_{j,k},K_{j,k},\kappa_{j,k},B
\bigr)
\end{equation*}
satisfying
\begin{align}
    \|(\phi_j^{-1})_{k,\rm NN}-(\phi_j^{-1})_{k}\|_{L^\infty(\phi_j(U_j))}
    \leq \varepsilon_{j,k}.
    \label{eq:epjk}
\end{align}

 The constants hidden in $O(\cdot)$ depend on $m$, $B:=\sup_{\bx\in \cM}\|\bx\|_{\infty}$, and the derivatives of the inverse coordinate maps $\phi_j^{-1}$'s  up to order $1$.

 After approximating the $(\phi_j^{-1})_{k}((\bz_j)_{1:d})$ by $(\phi_j^{-1})_{k,\rm NN}((\bz_j)_{1:d})$ in \eqref{eq:oracledecoder}, we will approximate the multiplication operation $(\phi_j^{-1})_{k,\rm NN}((\bz_j)_{1:d}) \times (\bz_j)_{d+1}$ by a neural network according to Lemma \ref{lemma:multiplication0}. 
These quantities satisfy $|(\bz_j)_{d+1}|\le 1$ and $ \sup_{(\bz_j)_{1:d} \in \phi_j(U_j)}|(\phi_j^{-1})_{k,\rm NN}((\bz_j)_{1:d})| \le B$ for all $j$ and $k$. 
 For any $\varepsilon_M >0$,
 the multiplication network for $\widetilde{\times}\left((\phi_j^{-1})_{k,\rm NN}((\bz_j)_{1:d}), (\bz_j)_{d+1}\right) $ such that

{\small
\begin{equation}
\label{eq:epM}
\resizebox{0.70\textwidth}{!}{$\displaystyle
\left|
\widetilde{\times}\left((\phi_j^{-1})_{k,\rm NN}((\bz_j)_{1:d}),(\bz_j)_{d+1}\right)
-
(\phi_j^{-1})_{k,\rm NN}((\bz_j)_{1:d})\times(\bz_j)_{d+1}
\right|
\le \varepsilon_M
$}
\end{equation}
}
 
 has $O(\log\varepsilon_M^{-1})$ layers. The width is bounded by $6$ and all parameters are bounded by $B$.

Combining \eqref{eq:epjk} and \eqref{eq:epM} gives rise to
\begin{equation*}
\resizebox{0.70\textwidth}{!}{$\displaystyle
\left|
\widetilde{\times}\left((\phi_j^{-1})_{k,\rm NN}((\bz_j)_{1:d}),(\bz_j)_{d+1}\right)
-
(\phi_j^{-1})_{k}((\bz_j)_{1:d})\times(\bz_j)_{d+1}
\right|
\le \varepsilon_M+\varepsilon_{j,k}
$}
\end{equation*}
for each $j$ and $k$. 
We set $\varepsilon_M/(2C_{\cM})$ and $\varepsilon_{j,k} = \varepsilon_D/(2C_{\cM})$, and then
\begin{equation*}
\resizebox{0.70\textwidth}{!}{$\displaystyle
\left|
\sum_{j=1}^{C_{\cM}}
\widetilde{\times}\left((\phi_j^{-1})_{k,\rm NN}((\bz_j)_{1:d}),(\bz_j)_{d+1}\right)
-
\sum_{j=1}^{C_{\cM}}
(\phi_j^{-1})_{k}((\bz_j)_{1:d})\times(\bz_j)_{d+1}
\right|
\le \varepsilon_D
$}
\end{equation*}
for all coordinate indexed by $k$.

 In summary, for any $\varepsilon_D \in (0,1)$, there is a network architecture $\cF_{\rm NN}(C_{\cM}(m+1),n,L_{D},p_{D},K_{D},\kappa_{D},B)$ with 
 \begin{equation}
 \begin{aligned}
     &L_{D}=O(\log \varepsilon_{D}^{-1}), \ p_{D}=O(C_{\cM}\varepsilon_{D}^{-m}), \\
    &  \ K_{D}=O(C_{\cM}\varepsilon_{D}^{-m}\log \varepsilon_{D}^{-1}), \ \kappa_{D}= O(\varepsilon_{D}^{-m})
     \label{eq:NND}
 \end{aligned}
 \end{equation}
 so that for the oracle decoder $\cD^*=\sum_{j=1}^{C_{\cM}} \phi_j^{-1}((\bz_j)_{1:d})\times (\bz_j)_{d+1}$ in \eqref{eq:oracledecoder}, there exists a network 
\begin{equation}
 \begin{aligned}
 & \cD_{\theta_d}= \sum_{j=1}^{C_{\cM}}\widetilde{\times}\left((\phi_j^{-1})_{k,\rm NN}((\bz_j)_{1:d}), (\bz_j)_{d+1}\right) \\
 &\in \cF_{\rm NN}(C_{\cM}(m+1),n,L_{D},p_{D},K_{D},\kappa_{D},B)
 \end{aligned}
 \end{equation}
 satisfying
 
{\small
\begin{equation}
\label{eq:approxd}
\resizebox{0.70\textwidth}{!}{$\displaystyle
\sup_{\{\bz_j\}_{j=1}^{C_\cM}: \, (\bz_j)_{1:d} \in \phi_j(U_j), \, (\bz_j)_{d+1} \in [0,1]}
\left\|
\cD_{\theta_d}(\{\bz_j\}_{j=1}^{C_\cM})
-
\cD^*(\{\bz_j\}_{j=1}^{C_\cM})
\right\|_\infty
\leq \varepsilon_D
$}
\end{equation}
}
 
 The constants hidden in $O(\cdot)$ in \eqref{eq:NND} depend on $m$, $B:=\sup_{\bu\in \cM}\|\bu\|_{\infty}$, and the derivatives of the inverse coordinate maps $\phi_j^{-1}$'s  up to order $1$.

\vspace{0.7em}
 \noindent $\bullet$ {\bf Propagator approximation theory.} We next present the neural network approximation result for the latent propagator. 
 
 We next apply Lemma \ref{lemma:Sobolev} for the approximation of $\cP^*(\cdot,\tau,\bzeta)$. For any $\varepsilon_P\in(0,1)$, there is a network architecture $\cF_{\rm NN}(C_{\cM}(m+1),C_{\cM}(m+1),L_P,p_P,K_P,\kappa_P,M_P)$ with 
 \begin{equation*}
 \begin{aligned}
     &L_P=O(\log \varepsilon_P^{-1}), \ p_P=O(\varepsilon_P^{-m}), \\
     & \ K_P=O(\varepsilon_P^{-m}\log \varepsilon_P^{-1}), \ \kappa_P= O(\varepsilon_P^{-m})
     \ M_P= O(1)
 \end{aligned}
 \end{equation*}
 so that there exists a network $\cP_{\theta_p}\in \cF_{\rm NN}(C_{\cM}(m+1),C_{\cM}(m+1),L_P,p_P,K_P,\kappa_P,M_P)$ satisfying
 \begin{align}
   \sup_{\bz(t,\bzeta) \in \cE^*(\cM)}  \|\cP_{\theta_p}(\bz(t,\bzeta),\tau,\bzeta)-\cP^*(\bz(t,\bzeta),\tau,\bzeta)\|\leq \varepsilon_P,
   \label{eq:approxp}
 \end{align}
 for all $t,\tau,\bzeta$.
 The constants hidden in $O(\cdot)$ depend on $C_{\cM},m$,  $\Lip_{\frakF}$, the derivatives of the coordinate maps $\phi_i$'s and $\phi_i^{-1}$'s and the partition of unity $\rho_i$'s up to order $1$.

\vspace{0.7em}
\noindent $\bullet$ {\bf Putting the approximation of encoder, decoder and latent propagator together.} Finally, we will put the neural network approximation result for the oracle encoder in \eqref{eq:approxe}, decoder in \eqref{eq:approxd} and latent propagator in \eqref{eq:approxp} together. With $\cE_{\theta_e}$ approximating $\cE^*$, $\cD_{\theta_d} $ approximating $\cD^*$, and $\cP_{\theta_p}$ approximating $\cP^*$, the evolutionary operator $\frakF$ can be approximated by $\mathfrak{F}_{{\tau}_{\theta}}$ as
$$ \mathfrak{F}_{{\tau}_{\theta}}(\bu(\bx,t,\bzeta))=\cD_{\theta_d}\circ\cP_{\theta_p}(\cE_{\theta_e}(\bu(\bx,t,\bzeta),\tau,\bzeta).$$
The approximation error of $\frakF$ is given as follows:
{ \small
\begin{align*}
 &\left\|\mathfrak{F}_{{\tau}_{\theta}}(\bu(\bx,t,\bzeta)) - \frakF(\bu(\bx,t,\bzeta))\right\|
 \\
  = & 
  \left\|\cD_{\theta_d}\circ\cP_{\theta_p}(\cE_{\theta_e}(\bu(\bx,t,\bzeta),\tau,\bzeta) - \cD^*\circ\cP^*(\cE^*(\bu(\bx,t,\bzeta),\tau,\bzeta)\right\|
  \\
  = 
  % term 1
  & 
  \left\|\cD_{\theta_d}\circ\cP_{\theta_p}(\cE_{\theta_e}(\bu(\bx,t,\bzeta),\tau,\bzeta)
  -\cD_{\theta_d}\circ\cP_{\theta_p}(\cE^*(\bu(\bx,t,\bzeta),\tau,\bzeta)\right\|
  \\
  % term 2
 &
  +\left\|\cD_{\theta_d}\circ\cP_{\theta_p}(\cE^*(\bu(\bx,t,\bzeta),\tau,\bzeta)-\cD_{\theta_d}\circ\cP^*(\cE^*(\bu(\bx,t,\bzeta),\tau,\bzeta)\right\|
  \\
  % term 3
  &+\left\|\cD_{\theta_d}\circ\cP^*(\cE^*(\bu(\bx,t,\bzeta),\tau,\bzeta) - \cD^*\circ\cP^*(\cE^*(\bu(\bx,t,\bzeta),\tau,\bzeta)\right\|
  %% upper bound lip
  \\
  \le 
  % term 1
  & 
  \Lip_{\cD_{\theta_d}} \Lip_{\cP_{\theta_p}}
  \left\|\cE_{\theta_e}(\bu(\bx,t,\bzeta)
  -\cE^*(\bu(\bx,t,\bzeta)\right\| \\
  % term 2
 & +\Lip_{\cD_{\theta_d}}\left\|\cP_{\theta_p}(\cE^*(\bu(\bx,t,\bzeta),\tau,\bzeta)-\cP^*(\cE^*(\bu(\bx,t,\bzeta),\tau,\bzeta)\right\|
  \\
  % term 3
  &+\left\|\cD_{\theta_d}\circ\cP^*(\cE^*(\bu(\bx,t,\bzeta),\tau,\bzeta) - \cD^*\circ\cP^*(\cE^*(\bu(\bx,t,\bzeta),\tau,\bzeta)\right\|
  %% upper bound error
  \\
  = 
  % term 1
  & 
  \Lip_{\cD_{\theta_d}} \Lip_{\cP_{\theta_p}}
  \varepsilon_E
  % term 2
  +\Lip_{\cD_{\theta_d}}\varepsilon_P
  % term 3
  +\varepsilon_D.
        \end{align*}}
  Setting $\varepsilon_E=\varepsilon$, $\varepsilon_D=\varepsilon$ and $\varepsilon_P=\varepsilon$ yields Theorem \ref{thm:nn}.

\section{Training dataset and evaluation strategy}
\label{appendix:data_split}

We construct a dataset of single-shot, multi-step forecast tuples of the form
{\small
\begin{equation}
\mathcal{S}
  :=
  \Big\{
      \big(\boldsymbol{u}(t_{kj},\zeta_k),\;
           \boldsymbol{u}(t_{kj}+\tau_{kji},\zeta_k)\big)
      : i=1,\ldots,I,\; j=1,\ldots,J
  \Big\}_{k=1}^K.
\end{equation}}

where $\boldsymbol{u}(t_{kj},\zeta_k) \in \mathbb{R}^n$ is the high-dimensional state, $\boldsymbol{\zeta_k}$ collects the physical parameters (e.g., Reynolds number), and $\tau$ is the forecast horizon. This representation supports direct learning of the evolution operator across both temporal and parametric variations.

The construction is similar of Dynamic Mode Decomposition (DMD)~\citep{schmid2022dynamic}. Unlike classical DMD, which typically uses consecutive snapshots $(\boldsymbol{u}(t), \boldsymbol{u}(t+\Delta t))$ and a linear map, we (i) allow arbitrary offsets $\tau$ and (ii) propagate in a nonlinear latent space conditioned on $\boldsymbol{\zeta}$. This generalization enables expressive modeling of nonlinear dynamics over a range of parameters and horizons.

In matrix form, the dataset can be organized as
\[
\mathcal{D} = 
\begin{bmatrix}
\boldsymbol{u}(t_{11}, \zeta_1) & \boldsymbol{u}(t_{11} + \tau_{111}, \zeta_1) & \tau_{111} & \boldsymbol{\zeta}_1 \\
\boldsymbol{u}(t_{11}, \zeta_1) & \boldsymbol{u}(t_{11} + \tau_{112}, \zeta_1) & \tau_{112} & \zeta_1 \\
\vdots & \vdots & \vdots & \vdots \\
\boldsymbol{u}(t_{1J}, \zeta_1) & \boldsymbol{u}(t_{1J} + \tau_{1JI}, \zeta_1) & \tau_{1JI} & \zeta_1 \\
\boldsymbol{u}(t_{21}, \zeta_2) & \boldsymbol{u}(t_{21} + \tau_{211}, \zeta_2) & \tau_{211} & \boldsymbol{\zeta}_2 \\
\vdots & \vdots & \vdots & \vdots \\
\boldsymbol{u}(t_{KJ}, \zeta_K) & \boldsymbol{u}(t_{KJ} + \tau_{KJI}, \zeta_K) & \tau_{KJI} & \zeta_K \\
\end{bmatrix}.
\]

\section{Latent dimension and training hyperparameters for 1D Burgers'}
\label{appendix:latent_dimension_choice}

All LEO variants use a two-dimensional latent space ($m=2$) for the Burgers' benchmark. To justify this choice, we estimate the intrinsic dimension of the dataset using the maximum-likelihood estimator of \cite{campadelli2015intrinsic}, implemented via the \texttt{scikit-dimension} package. Using $k=5$ and $k=10$ nearest neighbors, the estimated intrinsic dimension consistently yields values near $1.97$. This indicates that a two-dimensional latent space is both theoretically justified and empirically sufficient to capture the underlying solution manifold in a parsimonious manner. This observation is consistent with the theoretical framework in Section~\ref{sec:method}, where the complexity of the latent representation is governed by the intrinsic dimension of the solution manifold.

We tune the training hyperparameters of each LEO variant using Bayesian optimization with Weights \& Biases sweeps~\citep{wandb}. For each model, we run approximately $100$ configurations and retain the setting with the lowest validation mean-squared error under the $(\boldsymbol{\zeta},\tau)$ split described in Appendix~\ref{appendix:data_split}. Within the Burgers' benchmark, the encoder, decoder, and latent propagator architectures are fixed across sweeps; only scalar training hyperparameters are optimized.

For the 1D Burgers' benchmark, we consider four variants: direct, consistent, and hybrid LEO with an autoencoder backbone (D--LEO--AE, C--LEO--AE, H--LEO--AE), and direct LEO with a VAE backbone (D--LEO--VAE).
The resulting optimal hyperparameters are reported in Table~\ref{tab:Burgers'_hparams}. Here, $\lambda_{\mathrm{dir}}$ denotes the direct forecasting weight, $\lambda_{\mathrm{con,max}}$ denotes the maximum latent-consistency weight after warm-up, and the warm-up fraction controls the linear ramp of $\lambda_{\mathrm{con}}$ from $0$ to $\lambda_{\mathrm{con,max}}$. Entries marked ``--'' indicate quantities not used for a given variant.

\begin{table}[t]
  \centering
  \caption{Optimal training hyperparameters for Burgers' LEO variants (Batch size $B$, Learning rate $\alpha$, Number of epochs $N_e$, Direct loss $\lambda_{\mathrm{dir}}$, Consistency max $\lambda_{\mathrm{con,max}}$, Warm-up fraction $f_w$, KL weight $\beta$)}
  \label{tab:Burgers'_hparams}
  \setlength{\tabcolsep}{5.5pt}
  \begin{footnotesize}
  \begin{tabular}{lcccc}
    \toprule
      & D--LEO--AE
      & C--LEO--AE
      & H--LEO--AE
      & D--LEO--VAE\\
    \midrule
    $B$
      & 64
      & 64
      & 128
      & 64 \\
    $\alpha$
      & $9.88\times 10^{-4}$
      & $6.93\times 10^{-4}$
      & $9.41\times 10^{-4}$
      & $7.0\times 10^{-4}$\\
    $N_e$
      & 200
      & 75
      & 150
      & 100 \\
   $\lambda_{\mathrm{dir}}$
      & 0.514
      & --
      & 1.179
      & 1.70 \\
    $\lambda_{\mathrm{con,max}}$
      & --
      & 3.135
      & 0.671
      & -- \\
    $f_w$
      & --
      & 0.3768
      & 0.313
      & -- \\
   $\beta$
      & --
      & --
      & --
      & $1.2\times 10^{-5}$ \\
    \bottomrule
  \end{tabular}
  \end{footnotesize}
\end{table}

\section{Additional qualitative forecasts for Burgers' equation}
\label{appendix:Burgers'_qualitative}

\begin{figure}[t]
  \centering
  \includegraphics[width=\textwidth]{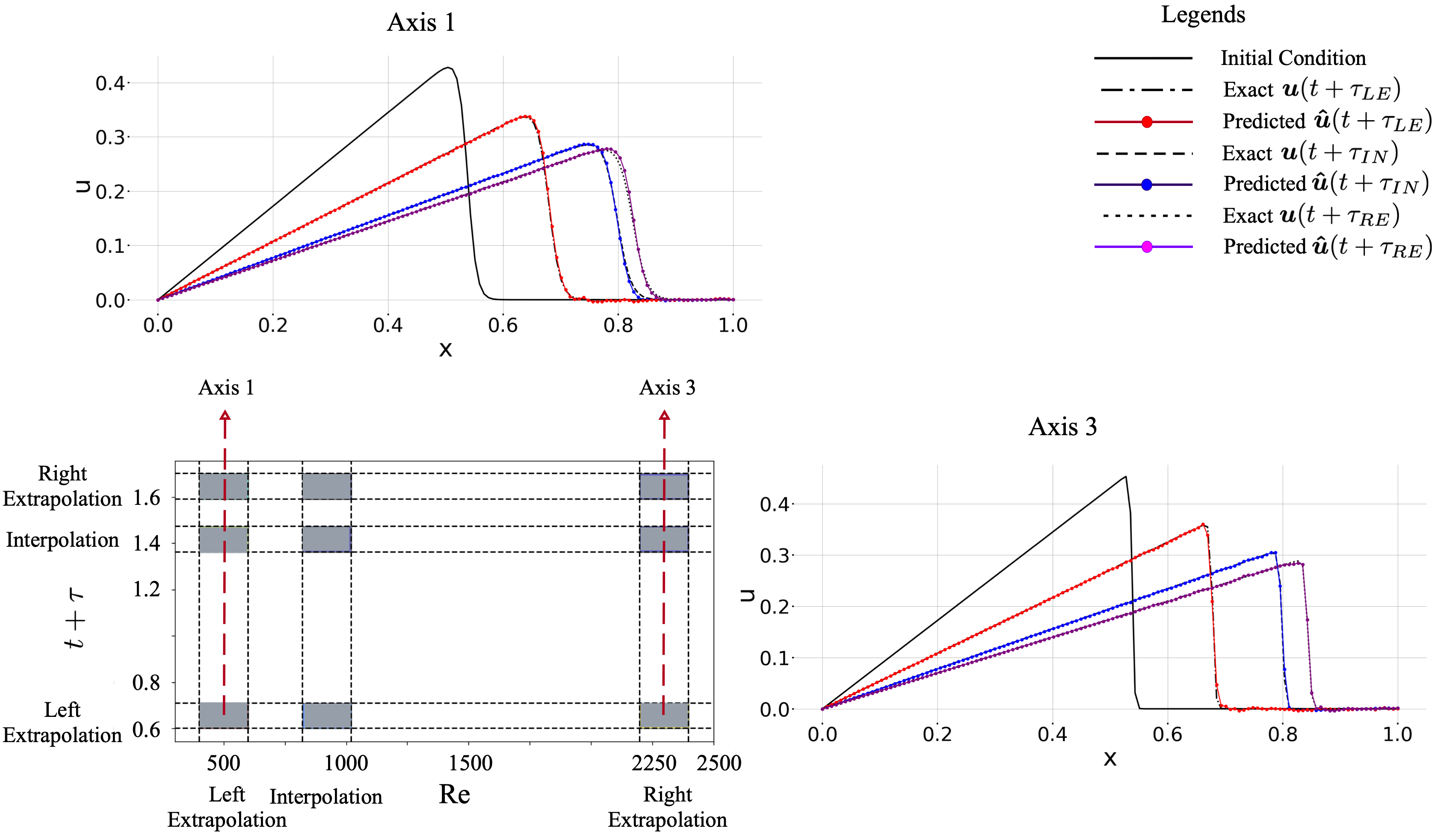}
  \caption{\textbf{Slice--wise Burgers' forecasts across the $(Re,\tau)$ grid.}
  The central panel shows the $3\times 3$ validation layout in $(Re,t+\tau)$ with grey rectangles indicating left/interpolation/right ranges in each
  variable. Two vertical slices are highlighted: \textbf{Axis~1} (low Reynolds) and \textbf{Axis~3} (high Reynolds). The top and right panels plot, for these slices, the initial condition
  $u(\cdot,t_0)$ (black), the exact solutions at three forecast horizons
  $\tau_{\mathrm{LE}}$, $\tau_{\mathrm{IN}}$, $\tau_{\mathrm{RE}}$ (dashed, dotted lines), and the corresponding Direct LEO (VAE) predictions (coloured markers).}
  \label{fig:Burgers'_axes}
\end{figure}

Figure~\ref{fig:Burgers'_axes} provides additional slice-wise forecasts of the Direct LEO model with a VAE backbone across the $(\mathrm{Re},\tau)$ grid used in Fig.~\ref{fig:Burgers'_mse}. The central panel reproduces the $3\times 3$ validation layout, with grey rectangles indicating the left/interpolation/right ranges in both $\mathrm{Re}$ and $t_a+\tau$.

The top and right panels show one-dimensional spatial profiles extracted along two representative slices of this grid. For fixed $\mathrm{Re}$ values on Axis~1 (low-Reynolds, diffusion-dominated regime) and Axis~3 (high-Reynolds, advection-dominated regime), we plot the anchor state $\bu(t_a)$ together with the exact solutions at three forecast horizons
$\tau_{\mathrm{LE}} < \tau_{\mathrm{IN}} < \tau_{\mathrm{RE}}$
and the corresponding Direct LEO predictions. Across both slices and all three horizons, the model accurately tracks the profile shape, peak location, and shock position, including cases in which $(\mathrm{Re},\tau)$ lies in left- or right-extrapolation zones. These qualitative results complement the zone-wise MSE reported in Fig.~\ref{fig:Burgers'_mse}, showing that the low quantitative errors correspond to stable long-horizon forecasts in both smooth, diffusion-dominated regimes and sharp, advection-dominated regimes.

\section{Positional--encoding latent propagator}
\label{appendix:pep_results}

\begin{figure}[t]
  \centering
  \includegraphics[width=\textwidth]{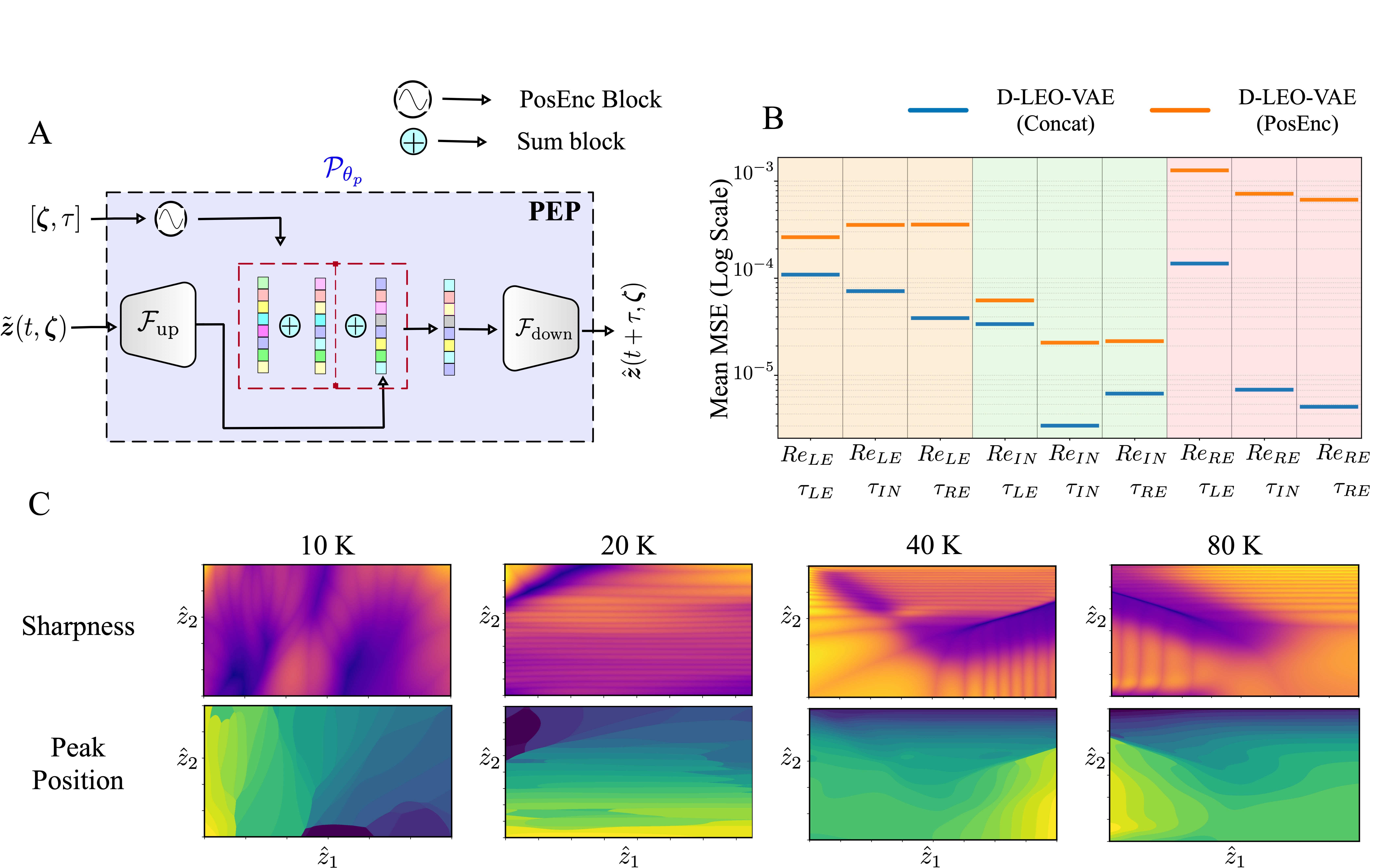}
  \caption{\textbf{Positional--encoding propagator (PEP) for Burgers'.}
  \textbf{(A)} Schematic of the positional--encoding propagator: the parameter pair $(\zeta,\tau)$ is mapped to a sinusoidal embedding and mixed with the latent code $\tilde z(t,\zeta)$ via up/down networks $\mathcal{F}_{\text{up}}$ and $\mathcal{F}_{\text{down}}$.
  \textbf{(B)} Zone--wise mean MSE (log scale) for D--LEO--VAE with concatenation (blue) versus positional encoding (orange) across the nine $(\mathrm{Re},\tau)$ validation zones. The concatenation--based propagator outperforms PEP in all zones. \textbf{(C)} Sharpness
  ($\max_x|\partial u/\partial x|$, top row) and peak position (bottom row) of decoded fields over a grid of latent points $(\hat z_1,\hat z_2)$ for training
  set sizes of 10k, 20k, 40k, and 80k tuples. Compared with the smooth, disentangled latent maps obtained with concatenation
  (Fig.~\ref{fig:scaling_disentangle}), the PEP latent structure is more irregular and entangled, especially at smaller data sizes.}
  \label{fig:pep_appendix}
\end{figure}

In addition to the concatenation-based latent propagator used in the main text, we experimented with a positional-encoding propagator (PEP) inspired by transformer architectures~\citep{vaswani2017attention}. For a given latent code $\tilde{\mathbf{z}}(t,\mathrm{Re})$ and input pair $(\mathrm{Re},\tau)$, PEP first maps $(\mathrm{Re},\tau)$ to a higher-dimensional sinusoidal embedding $\gamma(\mathrm{Re},\tau)$ and then mixes this embedding with the latent state using two small feed-forward networks $\mathcal{F}_{\mathrm{up}}$ and $\mathcal{F}_{\mathrm{down}}$, as sketched in Fig.~\ref{fig:pep_appendix}A. The encoder and decoder architectures are kept identical to those of the main D--LEO--VAE model; only the latent propagator is changed.

To compare the two parameterizations fairly, we train D--LEO--VAE with either concatenation (Concat) or positional encoding (PosEnc) on the same pool of $80{,}000$ training tuples
\[
\big(\bu(t_a,\mathrm{Re}),\,\bu(t_a+\tau,\mathrm{Re}),\,\tau,\,\mathrm{Re}\big),
\]
where $t_a$ is an anchor time and $\tau>0$ is a forecast horizon. Figure~\ref{fig:pep_appendix}B reports the mean MSE (log scale) for both variants across the nine $(\mathrm{Re},\tau)$ validation zones used in Fig.~\ref{fig:Burgers'_mse}. In every zone, the concatenation-based propagator attains lower error than its positional-encoding counterpart, sometimes by more than an order of magnitude, indicating that the additional positional embedding does not improve forecasting accuracy on this benchmark.

We also examine how the two propagators shape the latent space. For each model, we sample a uniform grid of points $(\hat z_1,\hat z_2)$ in the propagated latent plane, decode each point to a physical field, and compute two summary statistics from the decoded profiles: \emph{sharpness}, defined as $\max_x |\partial u/\partial x|$, which mainly reflects the effect of $\mathrm{Re}$, and \emph{peak position}, which is primarily governed by advective time shift. For the concatenation-based propagator, the corresponding plots in the main text (Fig.~\ref{fig:scaling_disentangle}) show a smooth, well-structured latent map: variation along one axis predominantly changes peak position, while variation along the other controls sharpness, and this disentanglement becomes clearer as the training set grows.

In contrast, Fig.~\ref{fig:pep_appendix}C shows that the PEP latent space is noticeably more irregular. The sharpness and peak-position fields exhibit striations, folds, and local discontinuities, particularly at smaller training sizes (10k--20k samples). This entangled geometry suggests that the positional-encoding propagator is more prone to overfitting and less effective at organizing latent directions into clean, physically interpretable factors of variation~\cite{bengio2013representation}. Together with the zone-wise MSE in Fig.~\ref{fig:pep_appendix}B, these results support our choice of the concatenation-based propagator as the default design in the main text. We emphasize, however, that this conclusion is drawn from the moderate-data regime and benchmark problems considered here: for substantially larger datasets or more complex systems, such as those targeted by emerging foundation models for PDEs, the relative advantage of positional encodings could change. Systematically exploring this regime is an interesting direction for future work, but lies beyond the scope of the present study.

\section{Decoder geometry around encoded and propagated latents}
\label{app:decoder_geometry}

We analyze the local geometry of the decoder $\mathcal{D}_{\theta_d}$ at two \emph{functionally equivalent} latent representations corresponding to the same forecasted Burgers' state: the encoded future
\[
\tilde{\boldsymbol{z}} = \mathcal{E}_{\theta_e}\big(\bu(t_0+\tau,\mathrm{Re})\big),
\]
and the propagated latent
\[
\hat{\boldsymbol{z}} = \mathcal{P}_{\theta_p}\big(\mathcal{E}_{\theta_e}(\bu(t_0,\mathrm{Re})),\tau,\mathrm{Re}\big),
\]
for a fixed $(t_0,\tau,\mathrm{Re})$ from the validation set. Although both latent codes decode to nearly identical high-dimensional states, i.e.,
\[
\mathcal{D}_{\theta_d}(\tilde{\boldsymbol{z}}) \approx \mathcal{D}_{\theta_d}(\hat{\boldsymbol{z}}),
\]
they occupy distinct regions of the latent space.

\begin{figure}[t]
  \centering
  \includegraphics[width=\columnwidth]{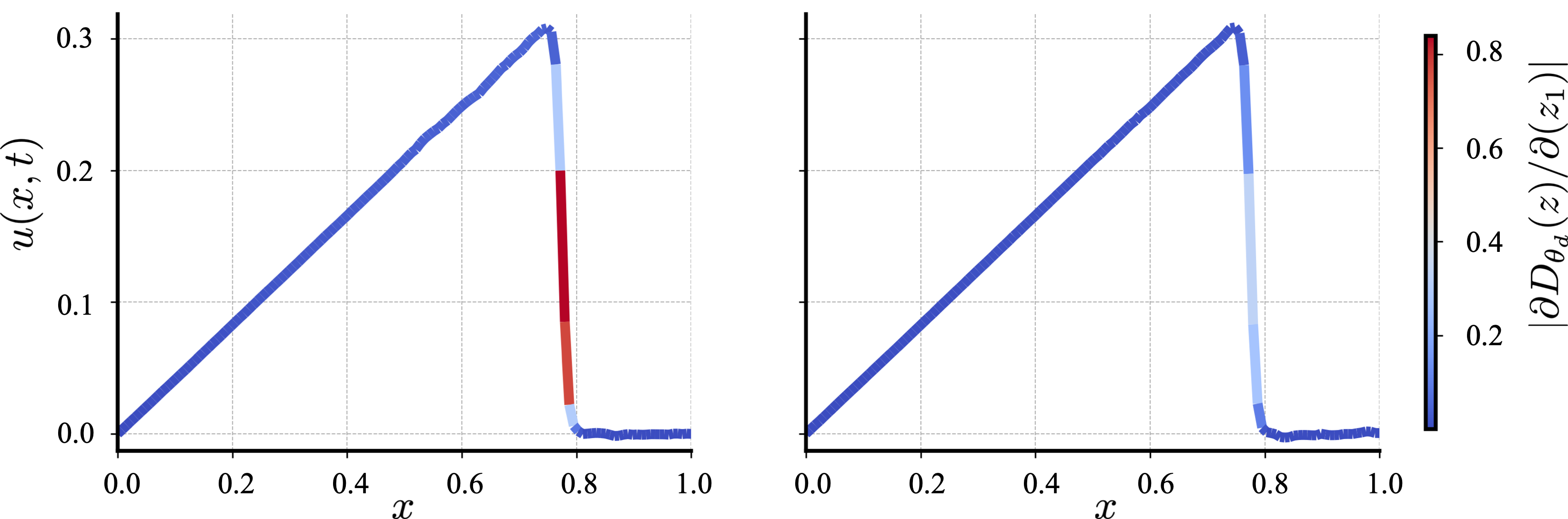}
  \caption{\textbf{Decoder sensitivity along the reconstructed profile.}
  Reconstructions of $u(x,t,\zeta)$ from the encoded latent $\tilde z$ (left) and propagated latent $\hat z$ (right), colour--coded by
  $|\partial \mathcal{D}_{\theta_d}(z)/\partial z_1|$.  Sensitivity is more concentrated and anisotropic near the shock at $\tilde z$ than at $\hat z$.}
  \label{fig:jacobian_colormap}
\end{figure}

Let
\[
J(\boldsymbol{z}) = \frac{\partial \mathcal{D}_{\theta_d}(\boldsymbol{z})}{\partial \boldsymbol{z}} \in \mathbb{R}^{n\times m}
\]
denote the decoder Jacobian. Figure~\ref{fig:jacobian_colormap} visualizes the decoded profiles, with each spatial location colored by the magnitude of the first Jacobian column
\[
\left|\frac{\partial \mathcal{D}_{\theta_d}(\boldsymbol{z})}{\partial z_1}\right|.
\]
At $\tilde{\boldsymbol{z}}$, the sensitivity is sharply localized near the shock, indicating strong and spatially concentrated dependence on latent perturbations. In contrast, at $\hat{\boldsymbol{z}}$, the sensitivity is more diffuse and uniformly distributed, reflecting a smoother local decoding map.

\begin{figure}[t]
  \centering
  \includegraphics[width=\columnwidth]{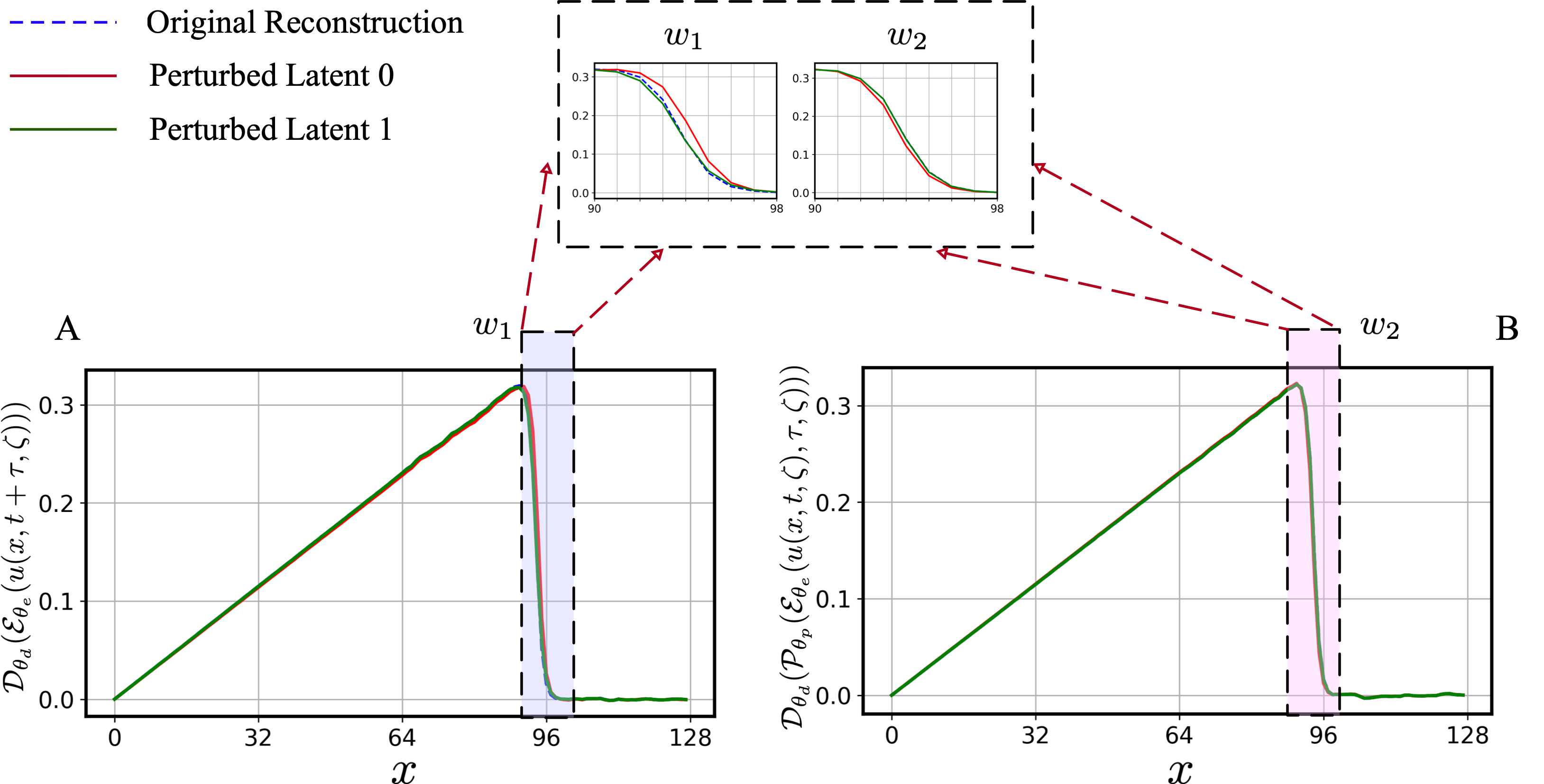}
  \caption{\textbf{Effect of small latent perturbations.}
  Decoder outputs for $\tilde z$ and $\hat z$ (blue) and for their perturbed versions $z + \varepsilon e_j$ along each latent axis (red/green). Insets zoom into the shock region, showing stronger sensitivity at
  $\tilde z$ and more stable behaviour at $\hat z$.}
  \label{fig:perturbation_output}
\end{figure}

To further probe local robustness, we consider perturbations along coordinate directions,
\[
\boldsymbol{z}_\varepsilon^{(j)} = \boldsymbol{z} + \varepsilon \mathbf{e}_j, \qquad \varepsilon = 0.01,
\]
and examine the decoded outputs. As shown in Fig.~\ref{fig:perturbation_output}, perturbations around $\tilde{\boldsymbol{z}}$ produce noticeable changes near the shock, whereas the same perturbations around $\hat{\boldsymbol{z}}$ lead to significantly smaller deviations. This indicates that $\hat{\boldsymbol{z}}$ lies in a flatter and more stable region of latent space.

\begin{figure}[t]
  \centering
  \includegraphics[width=\columnwidth]{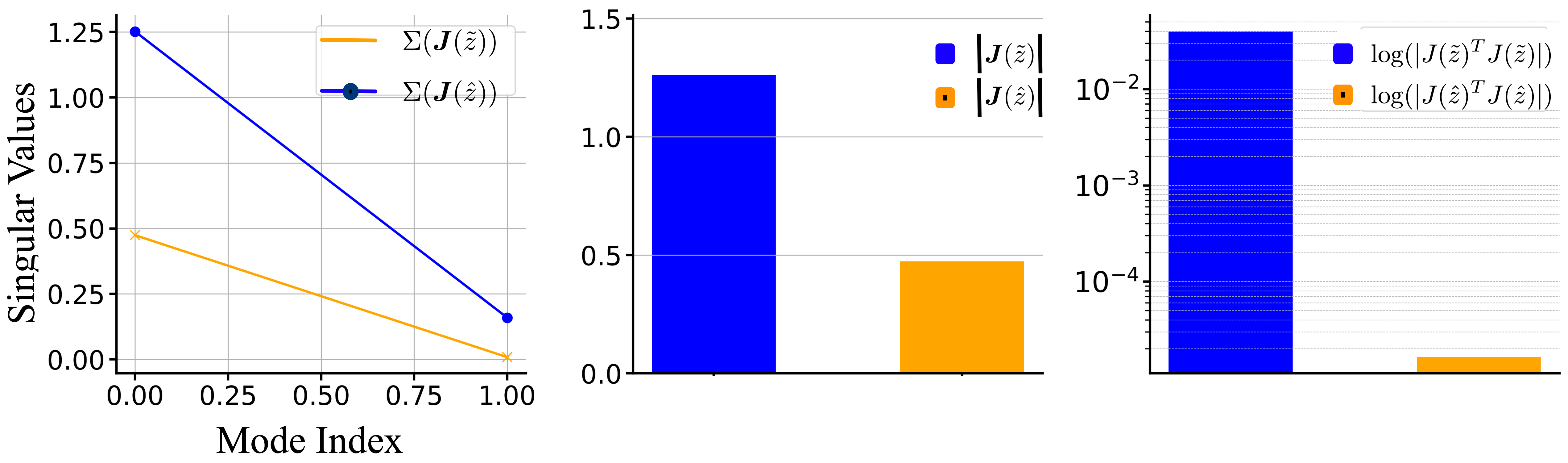}
  \caption{\textbf{Local geometry metrics of the decoder.}
  Singular values $\Sigma(J(z))$ (left), Frobenius norms $\|J(z)\|_F$ (middle), and $\log\det(J(z)^\top J(z))$ (right) at the encoded latent $\tilde z$ and propagated latent $\hat z$.  The smaller values at $\hat z$ indicate a better--conditioned, less expansive region of latent space.}
  \label{fig:decoder_geometry_metrics}
\end{figure}

We quantify these observations using spectral properties of the Jacobian. Writing
\[
J(\boldsymbol{z}) = U \Sigma V^\top,
\]
with singular values $\{\sigma_j(\boldsymbol{z})\}$, Fig.~\ref{fig:decoder_geometry_metrics} shows that $\tilde{\boldsymbol{z}}$ exhibits consistently larger singular values, larger Frobenius norm $\|J(\tilde{\boldsymbol{z}})\|_F$, and larger volume factor $\det(J(\tilde{\boldsymbol{z}})^\top J(\tilde{\boldsymbol{z}}))$ compared to $\hat{\boldsymbol{z}}$. These metrics indicate stronger local expansion and anisotropy near $\tilde{\boldsymbol{z}}$, whereas $\hat{\boldsymbol{z}}$ resides in a better-conditioned region with reduced sensitivity.

These results suggest that while multiple latent representations can yield the same decoded solution, the learned propagator preferentially maps states toward regions of latent space with smoother and more stable decoder geometry. This provides additional evidence that LEO implicitly regularizes the latent dynamics beyond mere prediction accuracy, favoring geometrically well-conditioned representations aligned with the underlying PDE evolution.

\section{Autoregressive baseline and runtime scaling}
\label{app:ae_lstm_runtime}

We compare LEO against a latent autoregressive baseline based on an autoencoder--LSTM (AE--LSTM) architecture. AE--LSTM first encodes the state into a two-dimensional latent space using the same encoder--decoder backbone as D--LEO--VAE, evolves the latent representation with a two-layer LSTM conditioned on $\mathrm{Re}$ in an autoregressive manner, and decodes the result back to the physical field. The architecture and parameter counts are matched to D--LEO--VAE to ensure a fair comparison.

For the Burgers' benchmark, AE--LSTM is trained on $\mathrm{Re}\in\{600,650,\dots,2225\}$ using 500 snapshots per simulation, with $t=0$ as the common anchor and the full temporal window used for training (i.e., no split over $\tau$). At test time, we fix the initial state and vary the forecast offset $\tau\in\{0.6,\dots,1.8\}$. D--LEO--VAE produces predictions in a single evaluation for each $\tau$, whereas AE--LSTM must sequentially propagate through all intermediate steps up to the target time.

\begin{figure}[t]
    \centering
    \includegraphics[width=\columnwidth]{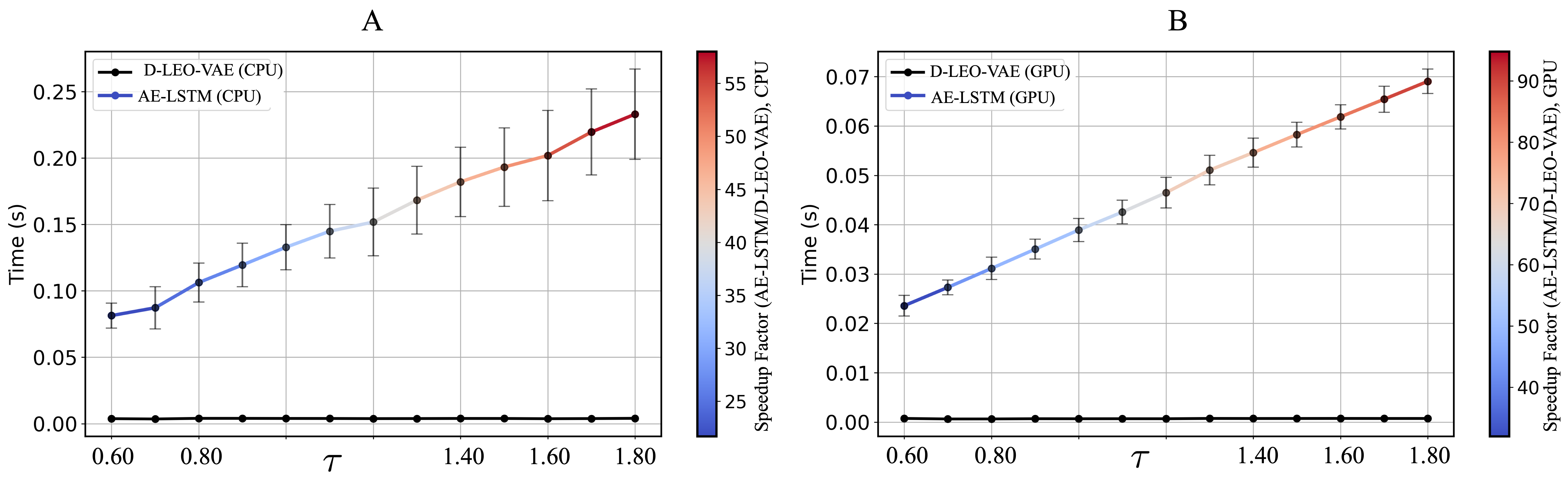}
    \caption{
        Prediction time comparison for D-LEO-VAE and AE--LSTM across temporal offsets \(\tau\) on CPU (A) and GPU (B).
        AE--LSTM runtime increases approximately linearly with \(\tau\) due to its autoregressive nature (\(\mathcal{O}(n)\)), whereas D-LEO-VAE achieves essentially constant cost (\(\mathcal{O}(1)\)). For large offsets, D-LEO-VAE attains up to \(50\times\) speedup on CPU and \(90\times\) on GPU.
    }
    \label{fig:cpu_gpu_comparison}
\end{figure}

Figure~\ref{fig:cpu_gpu_comparison} reports wall-clock prediction time on CPU and GPU, averaged over 300 runs per $\tau$. As expected, AE--LSTM exhibits approximately linear growth in runtime with $\tau$ due to its sequential $\mathcal{O}(\tau)$ rollout, while D--LEO--VAE maintains essentially constant $\mathcal{O}(1)$ cost across all forecast horizons. For the largest offsets, D--LEO--VAE achieves up to $50\times$ speedup on CPU and $90\times$ speedup on GPU, demonstrating substantial gains in latency and scalability.

\section{Model Architecture for 2D Advection Diffusion}
\label{app:2d_adv_diff_architecture}

We choose a latent dimension \(m=3\) to capture the essential structure of the
2D advection--diffusion dynamics while keeping the model parsimonious. This
reflects the increased spatial complexity relative to the 1D Burgers'' equation.
To justify this choice, we estimate the intrinsic dimensionality of the dataset
using the Maximum Likelihood Estimation (MLE) method with \(k=5\) and
\(k=10\) nearest neighbours (as implemented in the
\texttt{scikit-dimension} package~\citep{campadelli2015intrinsic}); both
settings yield an intrinsic dimension of approximately \(2.2\), supporting a
3D latent space.

Key hyperparameters are tuned via Bayesian optimisation to balance
reconstruction fidelity and long-horizon stability. The final configuration,
summarised in Table~\ref{tab:model_config2}, represents the best trade-off
between model complexity and performance. The model is trained on
36\,000 tuples, using the same D-LEO-VAE loss formulation as in the Burgers''
experiments.

\begin{table}[H]
\centering
\caption{Model configuration for the 2D advection--diffusion benchmark.}
\label{tab:model_config2}
\begin{tabular}{lc}
\hline
\textbf{Component}                       & \textbf{D-LEO-VAE} \\ \hline
Latent dimension (\(m\))                 & 3                  \\
Batch size (\(B\))                       & 64                 \\
Learning rate (\(\alpha\))              & \(7.0 \times 10^{-4}\) \\
Number of epochs                         & 75                 \\
KL-divergence weight (\(\beta\))        & \(1.152 \times 10^{-5}\) \\
Propagation loss weight (\(\lambda_{dir}\))    & 1.15               \\ 
Dataset size                             & 36\,000            \\
\hline
\end{tabular}
\end{table}

\section{Additional SWE dataset details and architecture}
\label{app:additional_swe_details}

The dataset is organised into train/validation/test splits. Within each split, simulations are grouped by random seeds, and each seed directory contains $100$ independent runs. Every run is stored as a NetCDF file with $88$ time steps on a $96\times 192$ latitude--longitude grid and a single vertical level, spanning roughly $20$ days of simulated time with output intervals of about $5$--$6$ hours. In our experiments we use the training split comprising $44$ seeds (i.e., $4400$ trajectories), and we focus on the zonal velocity component $u$ as the forecasted field.

To guide the choice of latent dimension for the SWE benchmark, we estimate the intrinsic dimension of the training snapshot cloud using the maximum likelihood estimator \citep{campadelli2015intrinsic}, implemented in the \texttt{scikit-dimension} package. Concretely, we flatten each snapshot $u(t)\in\mathbb{R}^{96\times 192}$ to $\mathbb{R}^{18432}$, draw random subsets of $10^4$--$5\times 10^4$ snapshots, standardize the data, and apply the MLE estimator with $k\in\{5,10,15,20\}$ nearest neighbours. Across subsamples and $k$ values, the intrinsic dimension concentrates around $\mathrm{id}\approx 18$, indicating a substantially low-dimensional manifold compared with the ambient space. Motivated by this estimate, all SWE experiments in this work use a $20$--dimensional latent space, providing a modest overparameterization relative to the inferred intrinsic dimension while remaining compact.

\section{Additional SWE forecasts from multiple anchor times}
\label{app:swe_multi_anchor}

\begin{figure}[H]
  \centering
  \includegraphics[width=\textwidth]{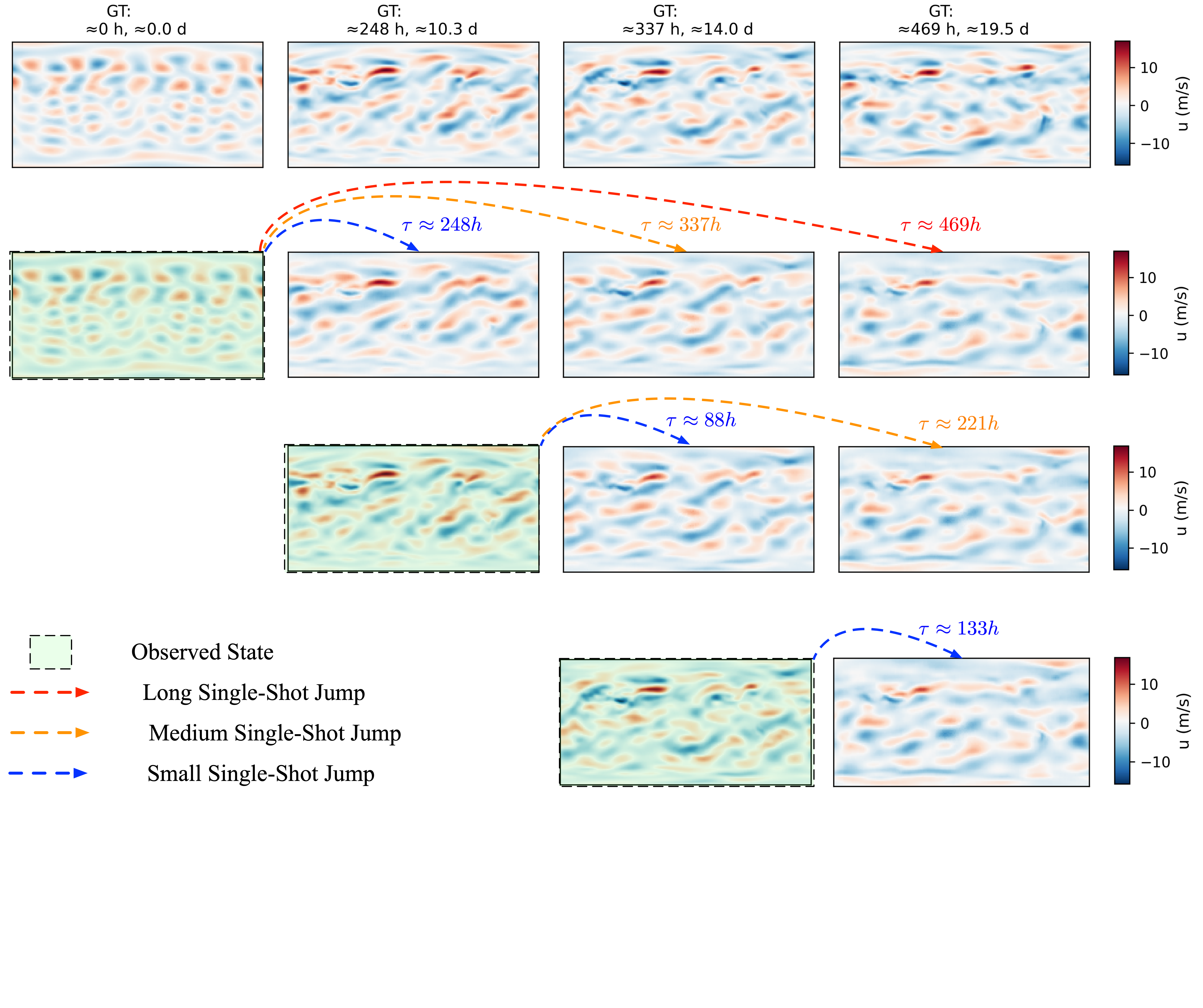}
  \caption{\textbf{Single--shot SWE forecasts from multiple observed states.}
  Top row: ground--truth zonal velocity snapshots at $\tau = 0,248h, 337h, 469h$ time
  steps. Rows 2--4: three observed states along the same trajectory (green dashed boxes) and LEO forecasts at later times. Dashed arrows are labeled with the corresponding temporal jump $\tau$ and colored by jump length (small, medium, long). In all cases the forecasts closely match the future snapshots qualitatively, illustrating that the latent evolution operator
  generalizes across anchor times and supports genuine state--to--state forecasting.}
  \label{fig:swe_multi_anchor}
\end{figure}

Figure~\ref{fig:swe_multi_anchor} complements the main shallow--water results by visualizing one--shot forecasts launched from multiple observation times along the \emph{same} test trajectory. For three anchors $t_a\in\{0,45,61\}$ (in model output steps), we use the observed zonal velocity $\bu(t_a)$ as input and predict the state at later times $t_a+\tau$ using the trained D--LEO--VAE. The dashed arrows in the figure indicate the jump lengths $\tau$ for each anchor.

Across all anchor times and jump lengths, the model reproduces the large--scale jets and vortices with similar fidelity, including long jumps up to $t=85$ (approximately $19.5$ days) starting from intermediate states.

\bibliographystyle{plainnat}
\bibliography{references}

\end{document}